\documentclass[conference]{IEEEtran}
\IEEEoverridecommandlockouts
\usepackage{cite}
\usepackage{amsmath,amssymb,amsfonts}
\usepackage{algorithmic}
\usepackage{graphicx}
\usepackage{textcomp}

\usepackage{xcolor}
\definecolor{lightgray_for_algorithm}{gray}{0.3}
\definecolor{Cerulean}{RGB}{0, 123, 167}
\definecolor{Red}{RGB}{255, 0, 0}
\definecolor{blue}{RGB}{0,0,0}
\definecolor{red}{RGB}{0,0,0}

\usepackage{algorithm}
\usepackage{subcaption, rotating}
\graphicspath{{./figs/}}
\usepackage{booktabs, array, multirow, makecell, colortbl}
\usepackage{bm, nicefrac}
\usepackage{hyperref}
\usepackage{cleveref}
\crefformat{equation}{Eq.~(#2#1#3)}
\crefformat{figure}{Fig.~#2#1#3}

\usepackage{amsthm}

\usepackage{enumitem}
\usepackage{bbold}

\newcommand{\tb}{\textbf}

\newcommand{\p}{$\,\pm\,$}
\newcommand{\ra}{{\color{Red}\alpha}}
\newcommand{\bn}{{\color{Cerulean}\nu}}

\newsavebox\CBox
\def\TB#1{\sbox\CBox{#1}\resizebox{\wd\CBox}{\ht\CBox}{\tb{#1}}}
\newcommand{\tabwidth}{1.49cm}
\newcommand{\secondtabwidth}{1.4cm}
\newcommand{\subfigwidth}{0.15}

\makeatletter 
\newcommand{\linebreakand}{%
  \end{@IEEEauthorhalign}
  \hfill\mbox{}\par
  \mbox{}\hfill\begin{@IEEEauthorhalign}
}
\makeatother 

\begin{document}

\title{Toward Evaluating Robustness of Reinforcement Learning with Adversarial Policy}


\author{
\IEEEauthorblockN{Xiang Zheng}
\IEEEauthorblockA{City University of Hong Kong\\
xzheng235-c@my.cityu.edu.hk}
\and
\IEEEauthorblockN{Xingjun Ma}
\IEEEauthorblockA{Fudan University\\
xingjunma@fudan.edu.cn}
\and
\IEEEauthorblockN{Shengjie Wang}
\IEEEauthorblockA{Tsinghua University\\
wangsj23@mails.tsinghua.edu.cn}
\and
\IEEEauthorblockN{Xinyu Wang}
\IEEEauthorblockA{Tencent Inc.\\
rainewang@tencent.com}
\linebreakand 
\IEEEauthorblockN{Chao Shen}
\IEEEauthorblockA{Xi'an Jiaotong University\\
chaoshen@mail.xjtu.edu.cn}
\and
\IEEEauthorblockN{Cong Wang\IEEEauthorrefmark{1}\thanks{\IEEEauthorrefmark{1}Corresponding author.}}
\IEEEauthorblockA{City University of Hong Kong\\
congwang@cityu.edu.hk}
}

\maketitle

\begin{abstract}
	Reinforcement learning agents are susceptible to evasion attacks during deployment. In single-agent environments, these attacks can occur through imperceptible perturbations injected into the inputs of the victim policy network. In multi-agent environments, an attacker can manipulate an adversarial opponent to influence the victim policy's observations indirectly. While adversarial policies offer a promising technique to craft such attacks, current methods are either sample-inefficient due to poor exploration strategies or require extra surrogate model training under the black-box assumption. To address these challenges, in this paper, we propose Intrinsically Motivated Adversarial Policy (IMAP) for efficient black-box adversarial policy learning in both single- and multi-agent environments. We formulate four types of adversarial intrinsic regularizers—maximizing the adversarial state coverage, policy coverage, risk, or divergence—to discover potential vulnerabilities of the victim policy in a principled way. We also present a novel bias-reduction method to balance the extrinsic objective and the adversarial intrinsic regularizers adaptively. Our experiments validate the effectiveness of the four types of adversarial intrinsic regularizers and the bias-reduction method in enhancing black-box adversarial policy learning across a variety of environments. Our IMAP successfully evades two types of defense methods, adversarial training and robust regularizer, decreasing the performance of the state-of-the-art robust WocaR-PPO agents by 34\%-54\% across four single-agent tasks. IMAP also achieves a state-of-the-art attacking success rate of 83.91\% in the multi-agent game YouShallNotPass. Our code is available at \url{https://github.com/x-zheng16/IMAP}.
\end{abstract}

\begin{IEEEkeywords}
Reinforcement learning, black-box evasion attack, adversarial policy, intrinsic motivation
\end{IEEEkeywords}

\section{Introduction}

\subsection{Background}

Reinforcement Learning (RL) agents are susceptible to a variety of attacks due to the vulnerabilities of their function approximators or policies themselves~\cite{zhang2021robust}. The growing application of RL agents in safety-critical systems, such as robotics and autonomous vehicles~\cite{huang2018adversarial,aradi2020survey,kiran2021deep,buddareddygari2022targeted}, underscores the need for the development of both certification methods~\cite{lutjens2020certified,everett2021neural,zhang2020robust,wu2021crop} and empirical evaluation methods~\cite{lin2017tactics,gleave2019adversarial,pinto2017robust,sun2020stealthy} to measure the robustness of deployed RL agents. Adversarial Policy (AP), a type of test-time evasion attack, has emerged as a crucial technique for assessing the robustness of the deployed RL engines or models~\cite{zhang2021robust,sun2021strongest,gleave2019adversarial,wu2021adversarial,guo2021adversarial,mo2022attacking}.

In single-agent environments, AP is developed to generate imperceptible perturbations on the inputs of the victim policy network. Sun et al.~\cite{sun2021strongest} proposed generating action perturbation via AP first and then crafting the corresponding state perturbation via the Fast Gradient Sign Method (FGSM). Mo et al.~\cite{mo2022attacking} suggested using two APs to select the attack timing and determine the worst-case victim action separately. Apart from these white-box methods, Zhang et al.~\cite{zhang2021robust} introduced SA-RL to learn the optimal black-box AP in dense-reward locomotion tasks. However, SA-RL requires knowledge of the victim policy's training-time rewards, which are difficult for the adversary to obtain under the black-box threat model.

In multi-agent competitive environments, AP is used to control an opponent agent to interact with the victim agent and indirectly influence the observation of the victim. Gleave et al.~\cite{gleave2019adversarial} first discovered this type of AP in two-player zero-sum competitive games, denoted as AP-MARL. Wu et al.~\cite{wu2021adversarial} suggested training an extra surrogate victim model by imitation learning first and then using an explainable Artificial Intelligent technique to identify the attack timing. Guo et al.~\cite{guo2021adversarial} developed AP learning for non-zero-sum games by simultaneously maximizing the adversary's and minimizing the victim's value functions. However, training an additional surrogate victim model yields only a marginal improvement in the attacking success rate~\cite{wu2021adversarial}. Moreover, all these AP learning methods are sample-inefficient due to their trivial dithering exploration methods.

\subsection{Motivations and Design Rationale}

\paragraph{Motivations} In this work, we explore and propose Intrinsically Motivated Adversarial Policy (IMAP) for efficient black-box AP learning in both single- and multi-agent environments. There are three main challenges.
Firstly, efficient exploration is known to be critical for RL algorithms to improve performance and reduce sample complexity. However, existing AP learning methods all suffer from poor exploration in both single- and multi-agent environments as they all explore in an ad-hoc and trivial manner by heuristically perturbing the outputs of the AP with Gaussian noise. To address this, we design four types of adversarial intrinsic regularizers to enhance the exploration of the AP in a principled way. Adversarial intrinsic regularizers encourage the AP to explore novel states more efficiently so as to uncover potential vulnerabilities of the black-box victim policy.
Secondly, the incorporation of adversarial intrinsic regularizers presents a new challenge: how to effectively balance the original extrinsic objective and the newly introduced adversarial intrinsic regularizers. To simplify the hyperparameter search for the optimal temperature parameter that controls the strength of the regularization, we employ constrained policy optimization to develop an adaptive balancing strategy.
Thirdly, existing AP methods, except for AP-MARL, all follow a relaxed black-box threat model or require extra surrogate victim model training. One of the key assumptions on the knowledge of the adversary made by AP-MARL is that the adversary against the deployed victim policy does not have access to the training-time rewards and the value function of the deployed victim agent. To address this, we stick to the (unrelaxed) black-box assumptions on the knowledge of the adversary to design our IMAP and do not rely on extra surrogate victim models.

\paragraph{Design Rationale} To encourage the exploration of the AP, we design four types of adversarial intrinsic regularizers for IMAP that maximize the adversarial State Coverage (SC), Policy Coverage (PC), Risk (R), and Divergence (D).
All four types of adversarial intrinsic regularizers are designed for the AP to uncover the potential vulnerabilities of the victim policy efficiently and have solid theoretical support, including state entropy~\cite{hazan2019provably}, policy cover~\cite{agarwal2020pc}, constrained policy optimization~\cite{tessler2018reward}, and policy diversity~\cite{hong2018diversity,flet2021adversarially}.
Intuitively, efficient exploration for black-box AP learning can involve either uniform state visitation (maximizing the adversarial SC) or maximizing deviation from explored regions (maximizing the adversarial PC). Further, the R- and D-driven adversarial intrinsic regularizers are also well-motivated, with the former encouraging the AP to lure the victim policy into adversarial states and the latter encouraging the AP to keep deviating from its past policies to be diverse. In addition to promoting the exploration of the AP, the inductive bias introduced by adversarial intrinsic regularizers may distract the adversary from its objective—decreasing the performance of the victim policy—in the final stage of AP learning. We find that such a distraction phenomenon exists in sparse-reward tasks and design a novel bias-reduction method to enhance the performance of IMAP further.

\begin{figure}[t]
	\centering
	\begin{subcaptionblock}{\columnwidth}
		\centering
		\includegraphics[width=\subfigwidth\columnwidth]{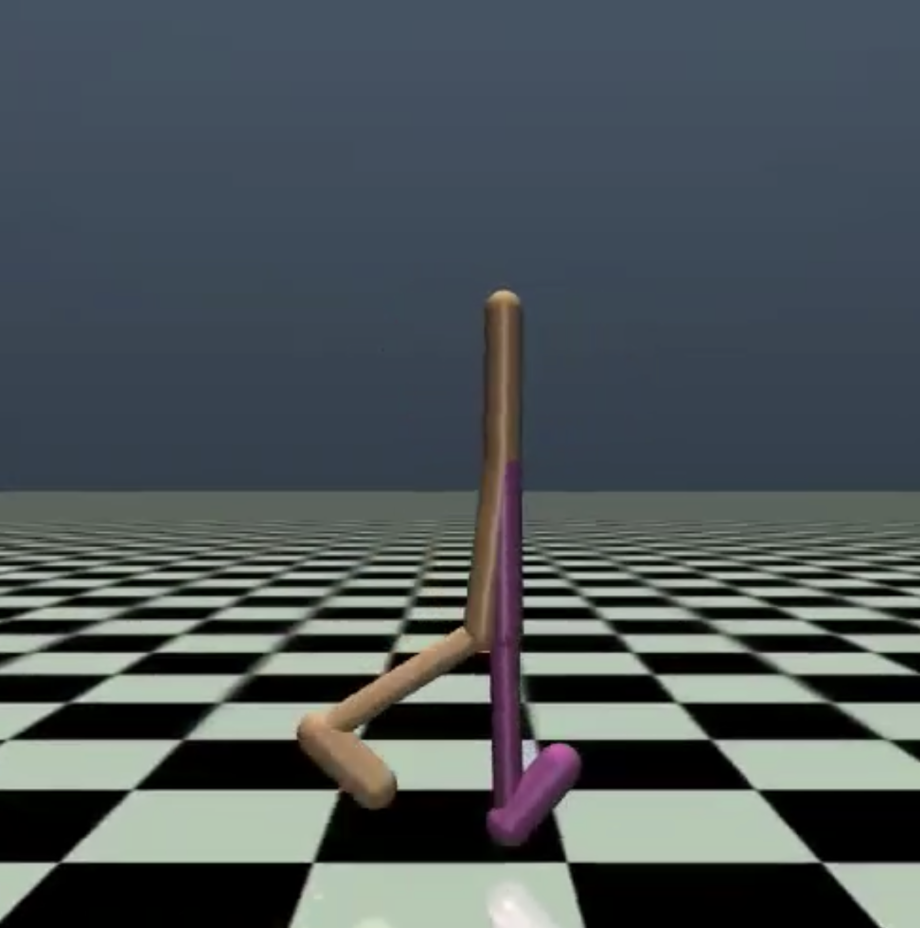}
		\includegraphics[width=\subfigwidth\columnwidth]{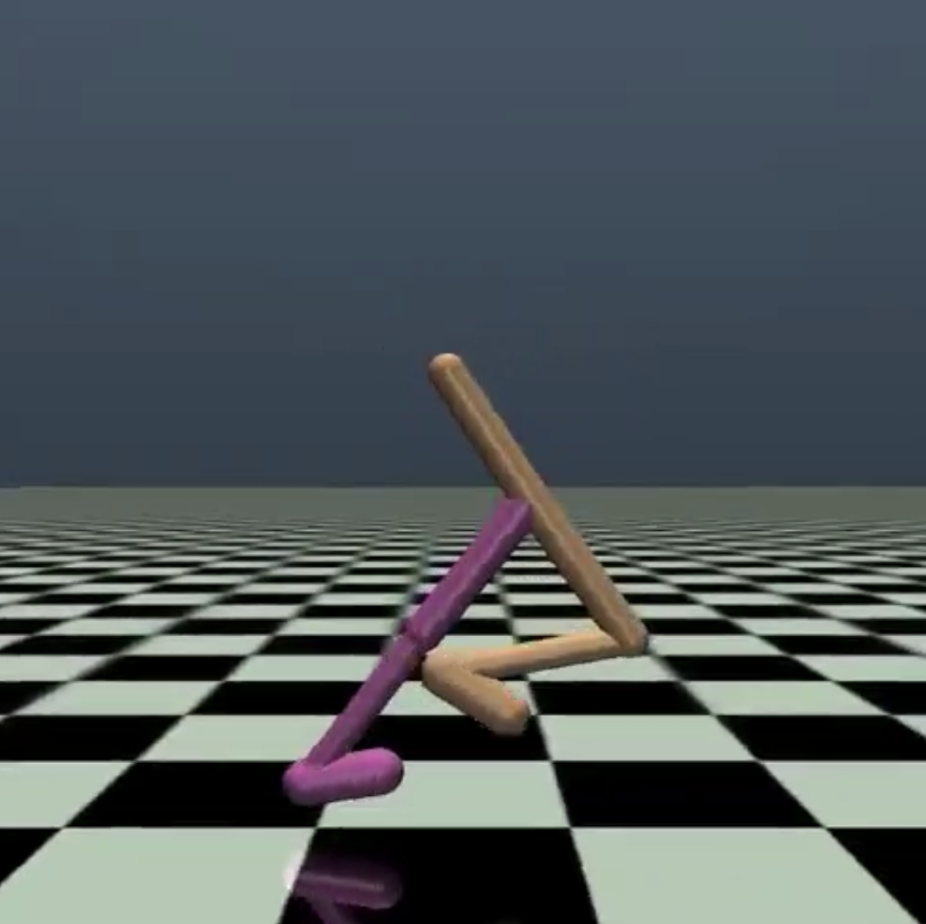}
		\includegraphics[width=\subfigwidth\columnwidth]{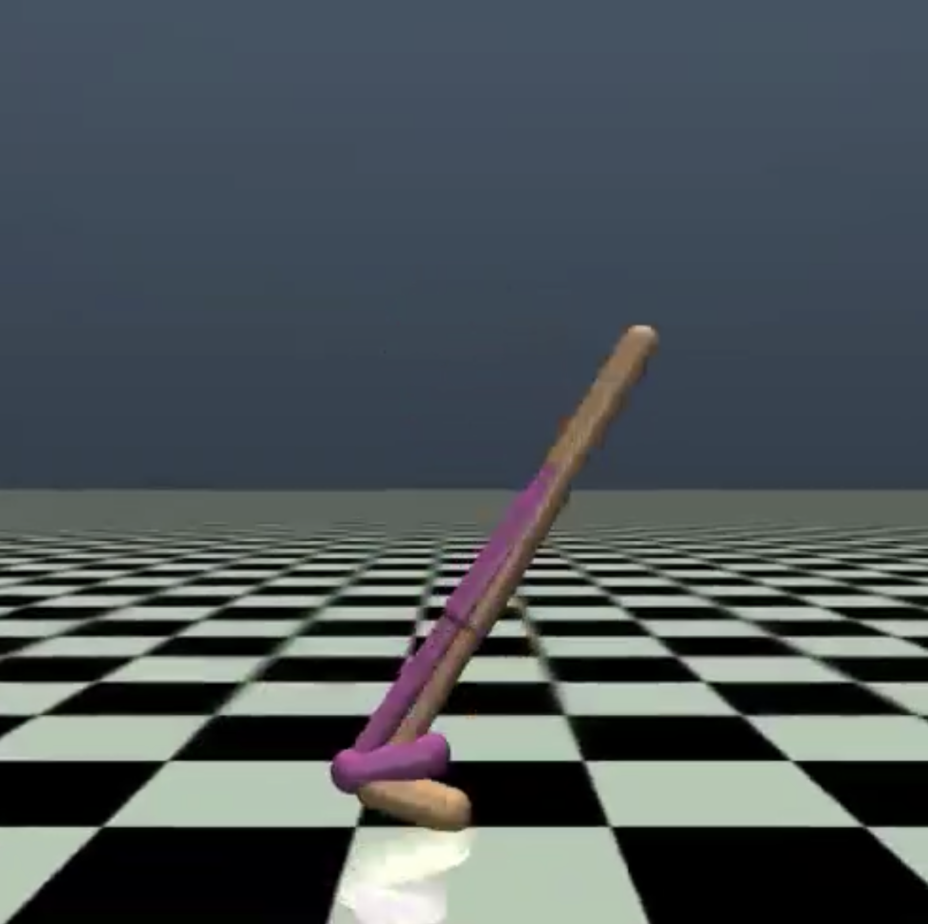}
		\includegraphics[width=\subfigwidth\columnwidth]{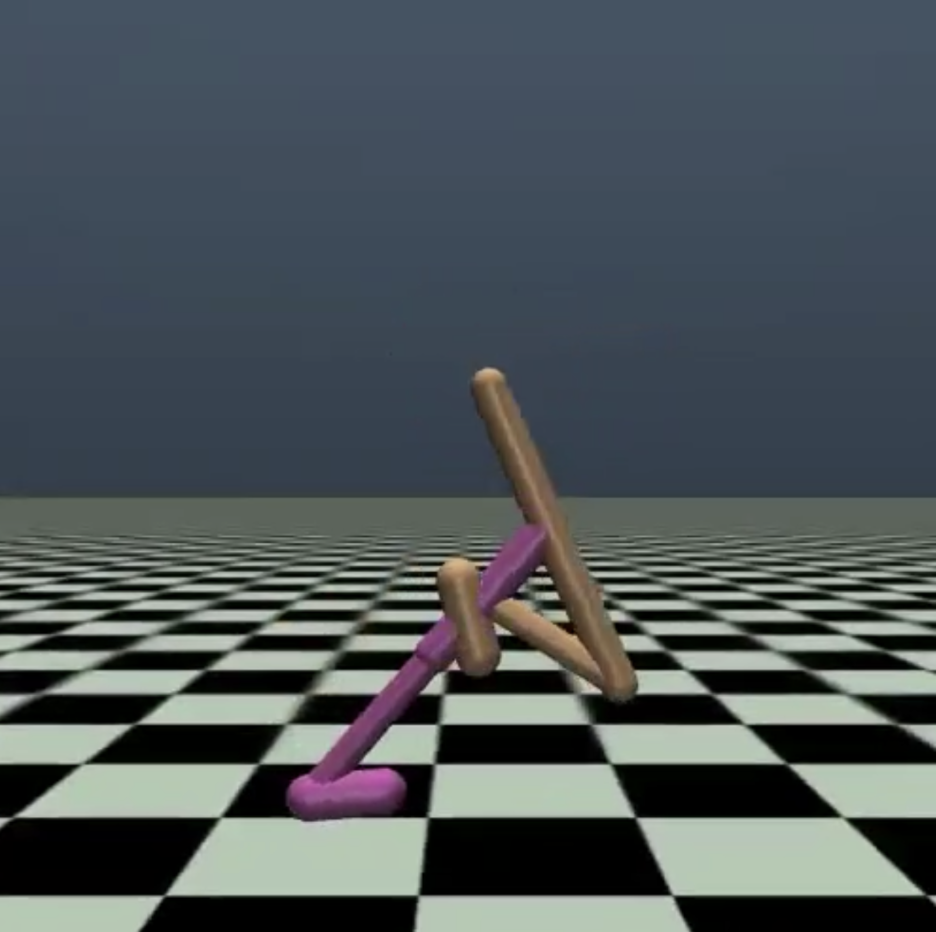}
		\includegraphics[width=\subfigwidth\columnwidth]{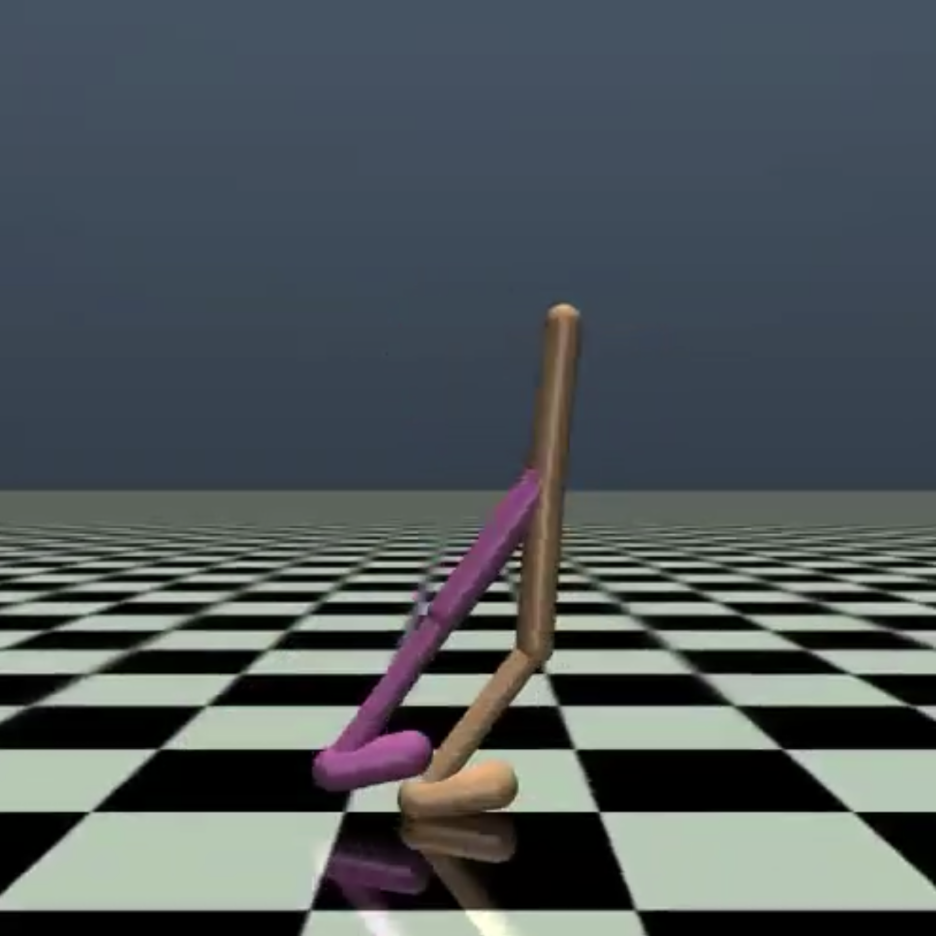}
		\includegraphics[width=\subfigwidth\columnwidth]{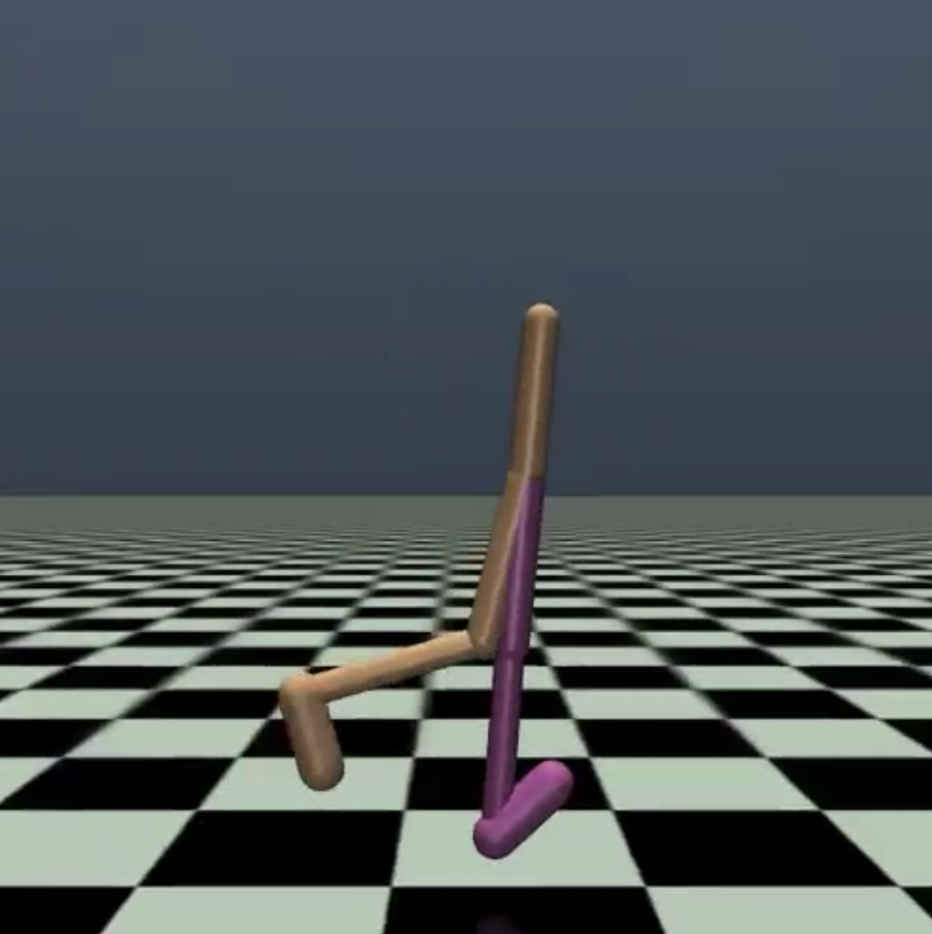}
		\phantomcaption{SA-RL: fail to make the robust agent fall or slow down}\label{fig:sa-attack1}
	\end{subcaptionblock}
	\begin{subcaptionblock}{\columnwidth}
		\vspace{3mm}
		\centering
		\includegraphics[width=\subfigwidth\columnwidth]{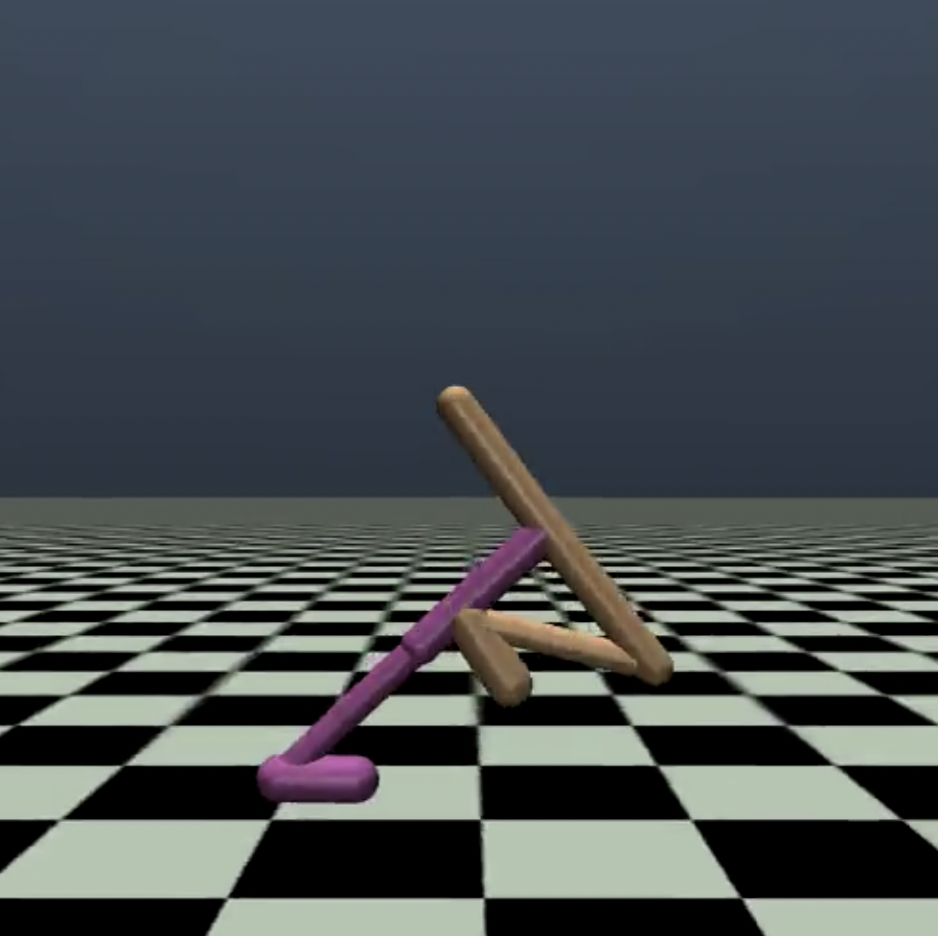}
		\includegraphics[width=\subfigwidth\columnwidth]{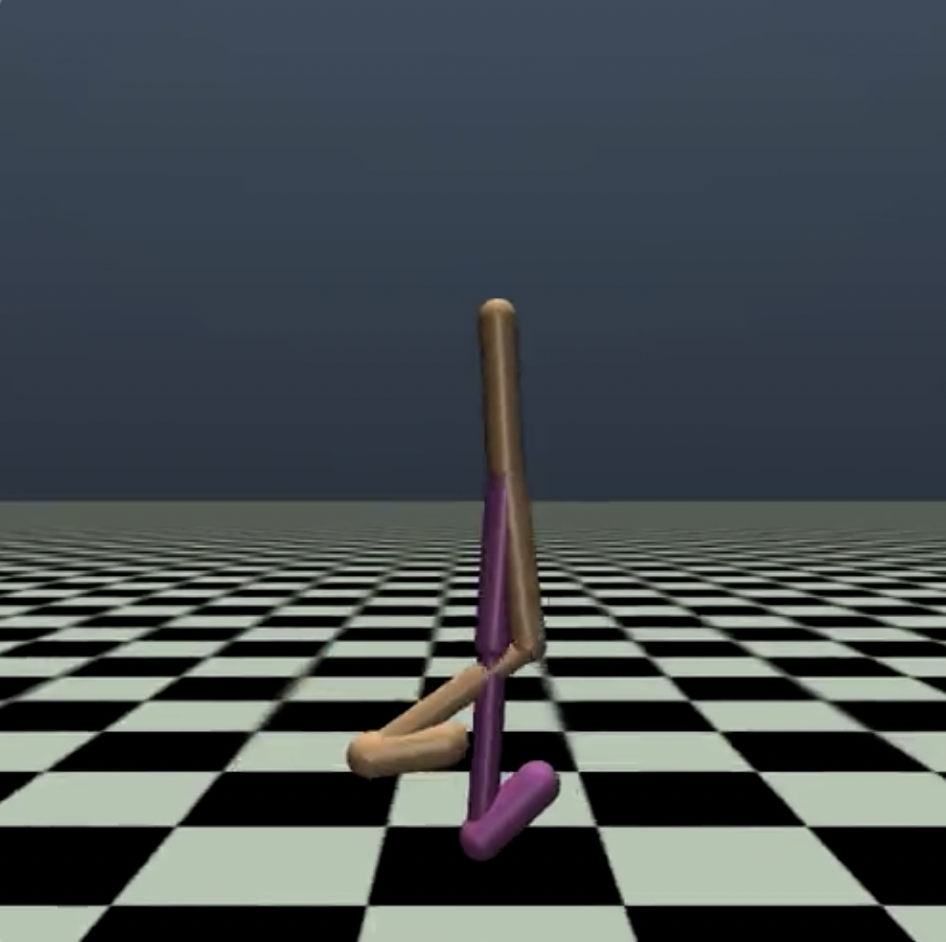}
		\includegraphics[width=\subfigwidth\columnwidth]{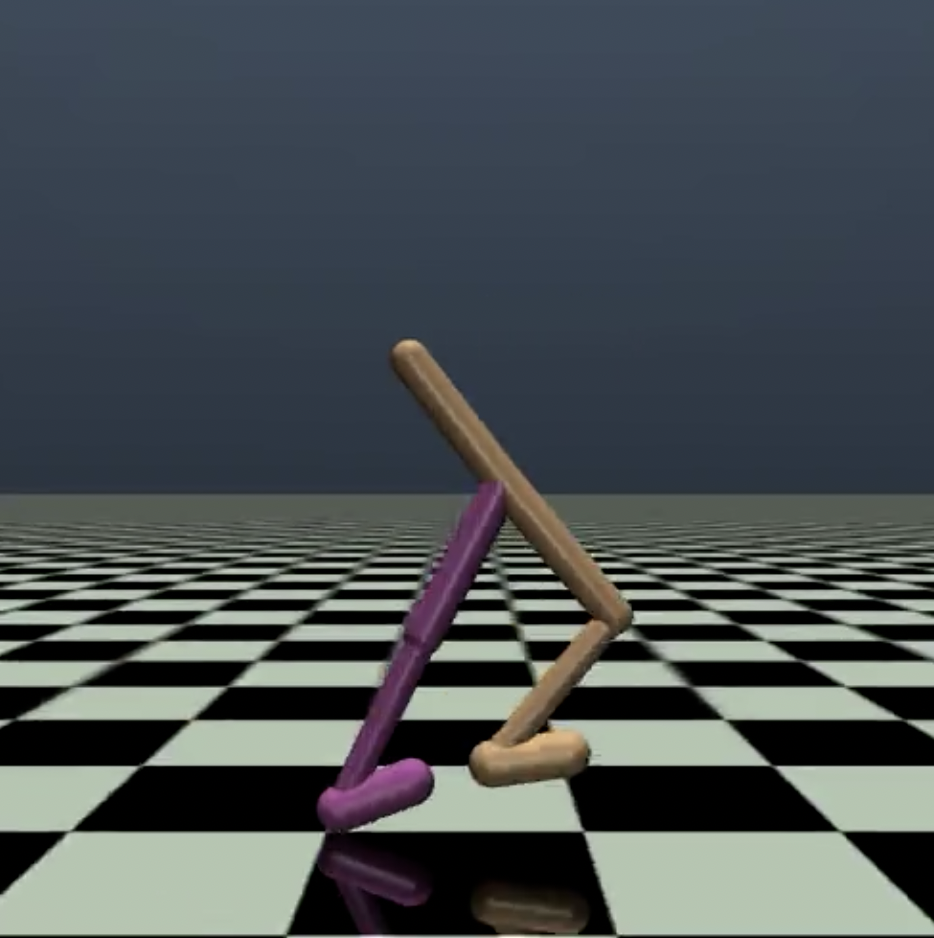}
		\includegraphics[width=\subfigwidth\columnwidth]{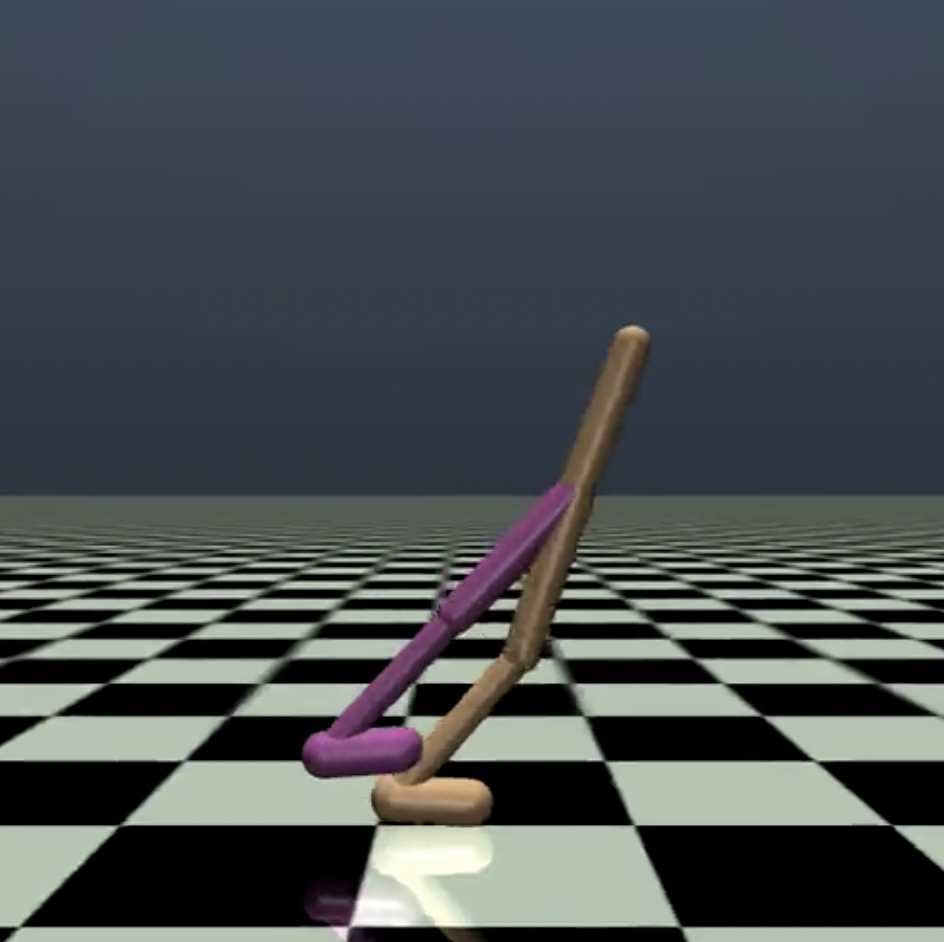}
		\includegraphics[width=\subfigwidth\columnwidth]{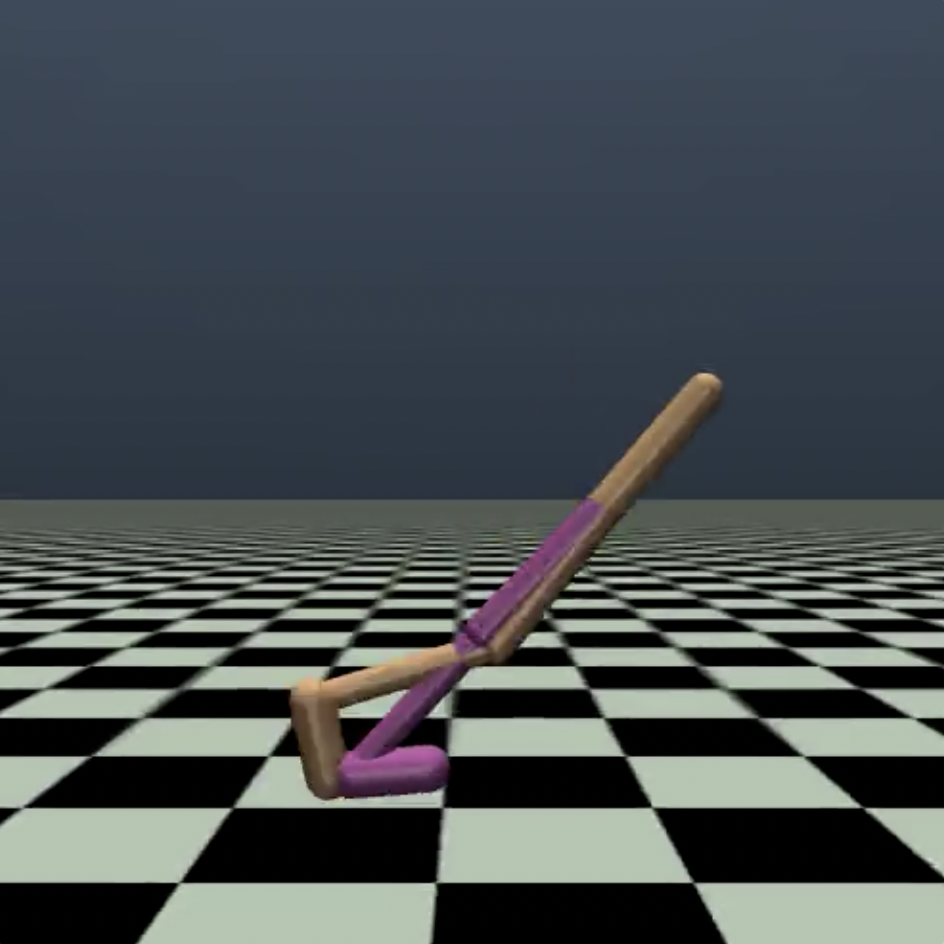}
		\includegraphics[width=\subfigwidth\columnwidth]{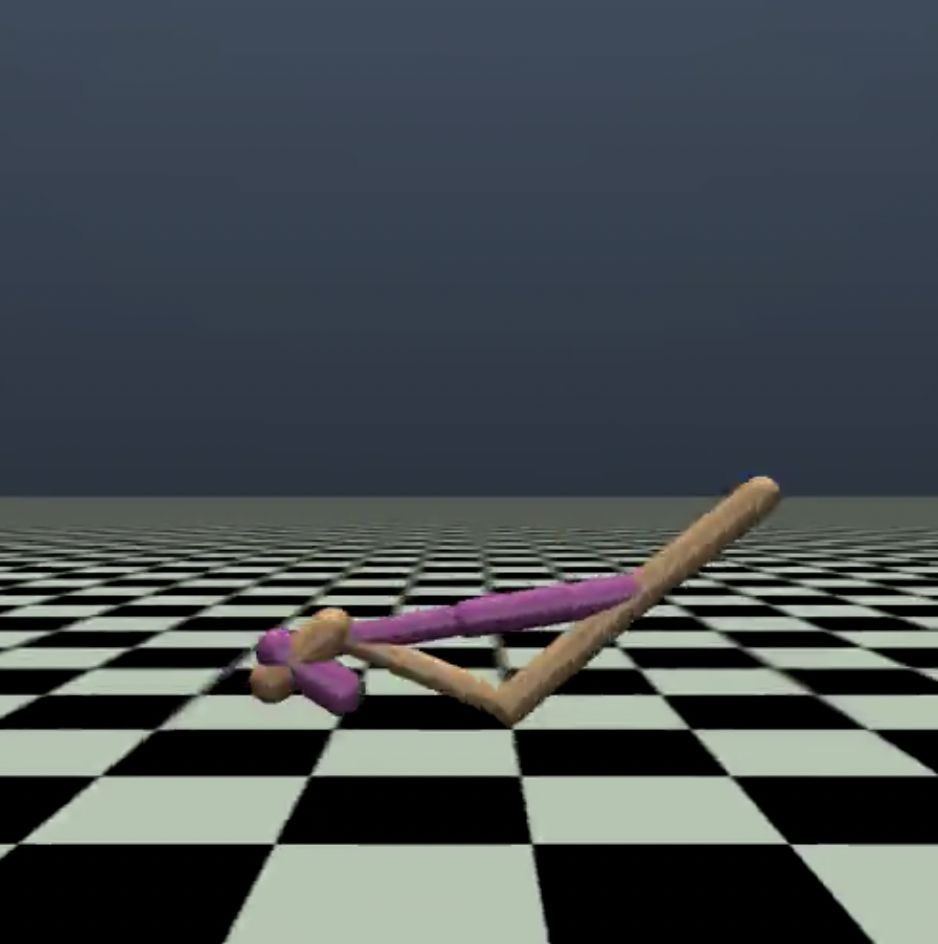}
		\phantomcaption{IMAP (ours): successfully make the robust agent fall}\label{fig:sa-attack2}
	\end{subcaptionblock}
	\caption{The robust victim agent—trained with the state-of-the-art defense method WocaR~\cite{liang2022efficient}—is attacked by (\tb{top}) the state-of-the-art AP method SA-RL and (\tb{bottom}) our IMAP in the single-agent environment Walker. Though the WocaR Walker learned to lower its body to be robust, our IMAP can find its vulnerable states and successfully lure the victim to lean forward and fall.}
	\label{fig: walker2d-IMAP}
\end{figure}

\begin{figure}[t]
	\centering
	\begin{subcaptionblock}{\columnwidth}
		\centering
		\includegraphics[width=\subfigwidth\columnwidth]{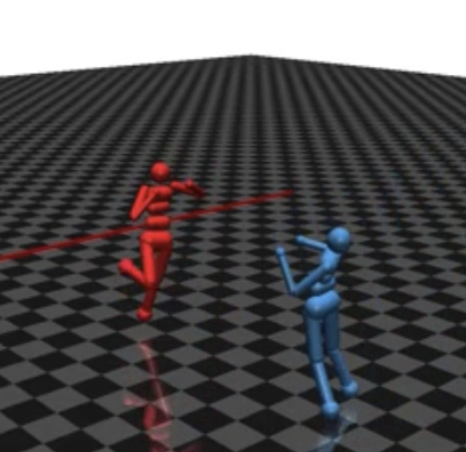}
		\includegraphics[width=\subfigwidth\columnwidth]{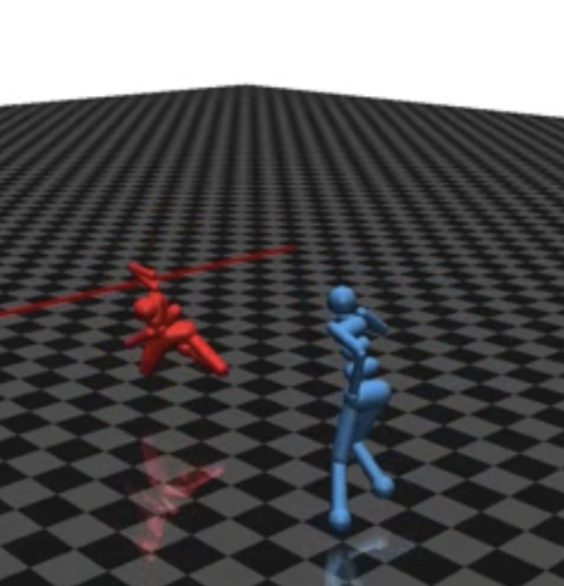}
		\includegraphics[width=\subfigwidth\columnwidth]{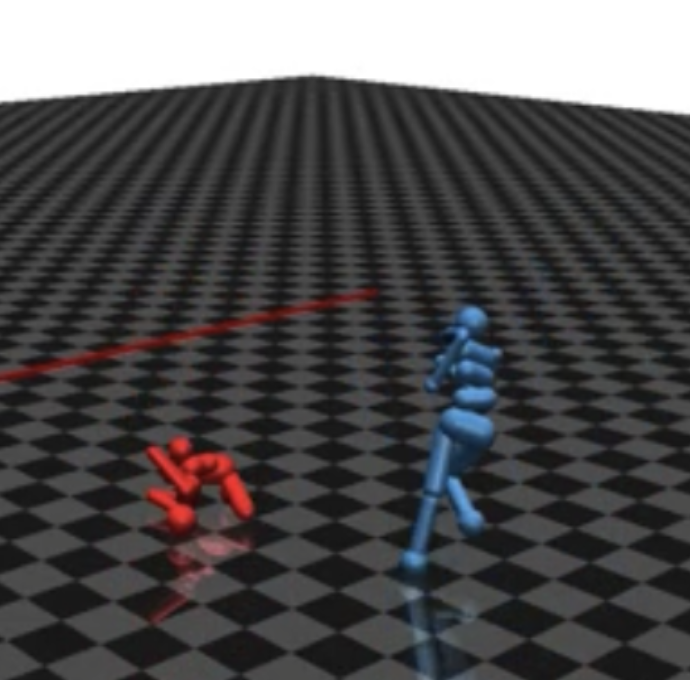}
		\includegraphics[width=\subfigwidth\columnwidth]{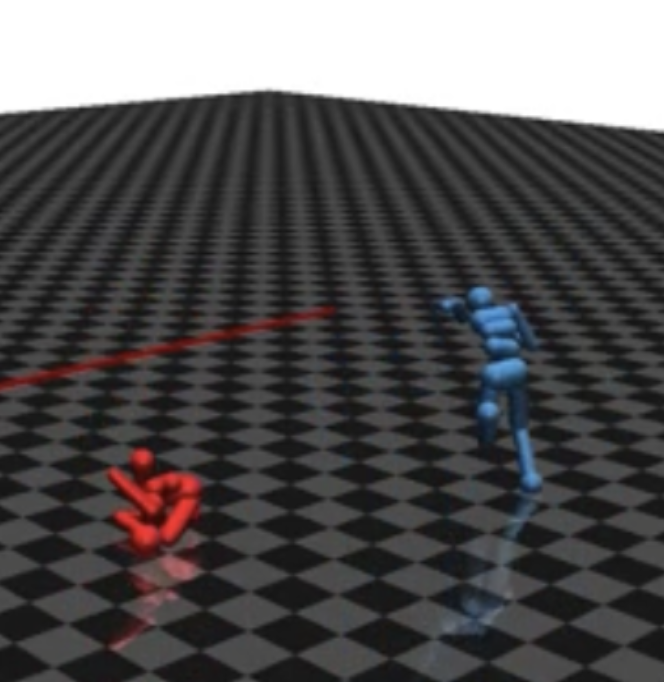}
		\includegraphics[width=\subfigwidth\columnwidth]{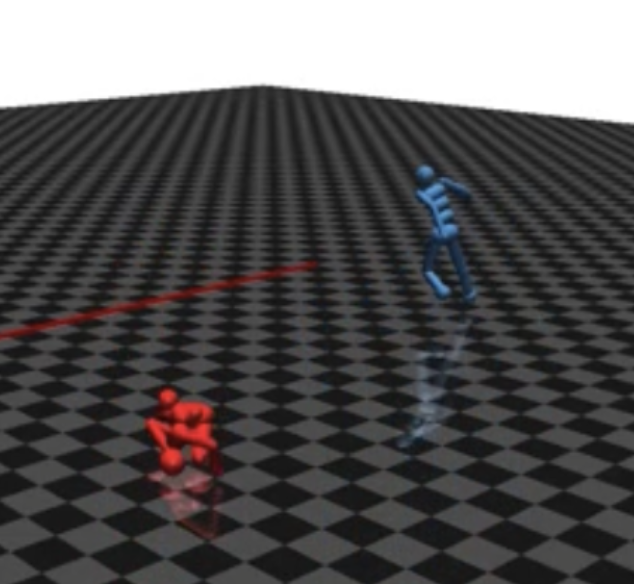}
		\includegraphics[width=\subfigwidth\columnwidth]{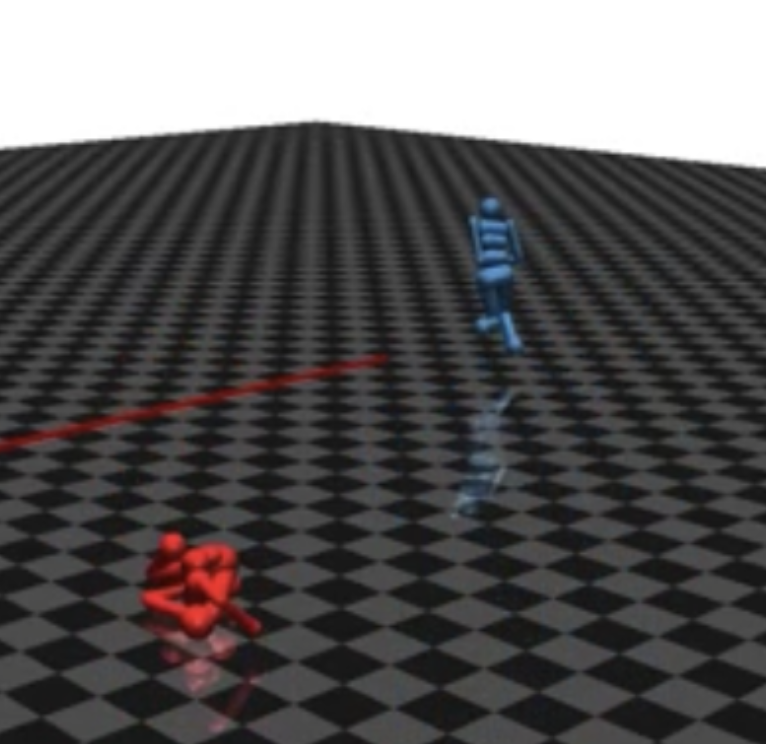}
		\phantomcaption{AP-MARL: fail to cause the victim to take poor actions}\label{fig:ma-attack1}
	\end{subcaptionblock}
	\begin{subcaptionblock}{\columnwidth}
		\vspace{3mm}
		\centering
		\includegraphics[width=\subfigwidth\columnwidth]{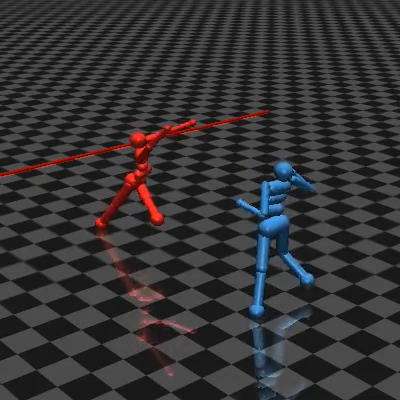}
		\includegraphics[width=\subfigwidth\columnwidth]{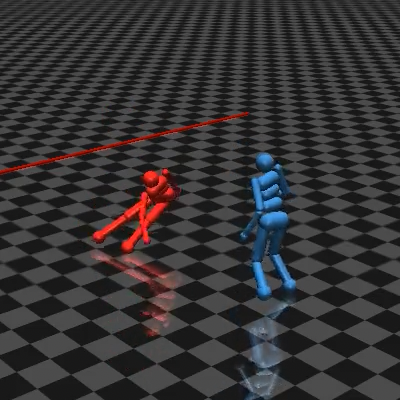}
		\includegraphics[width=\subfigwidth\columnwidth]{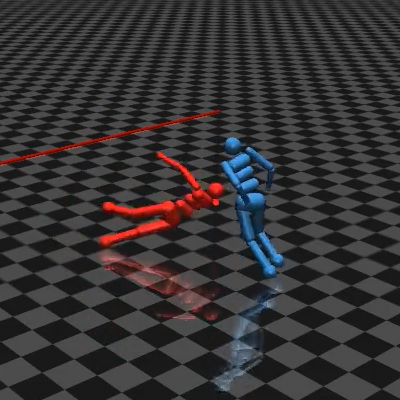}
		\includegraphics[width=\subfigwidth\columnwidth]{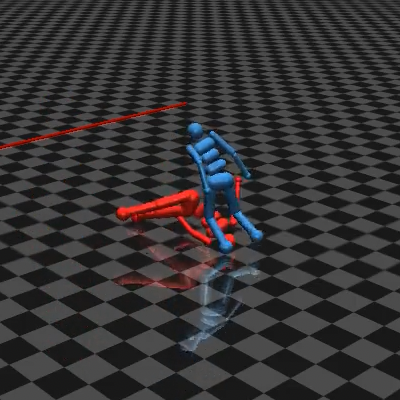}
		\includegraphics[width=\subfigwidth\columnwidth]{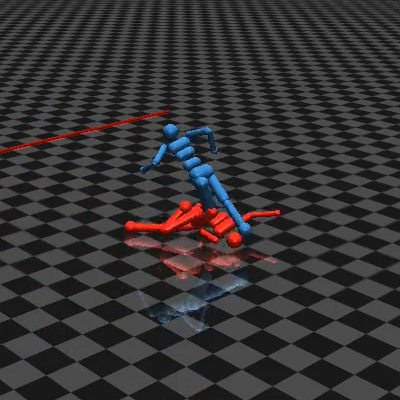}
		\includegraphics[width=\subfigwidth\columnwidth]{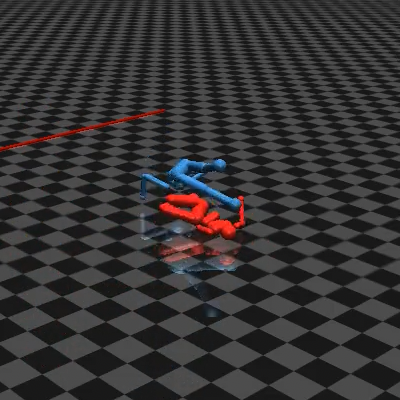}
		\phantomcaption{IMAP (ours): successfully block the victim and make it fall}\label{fig:ma-attack3}
	\end{subcaptionblock}
	\caption{The victim (in blue) is attacked by an adversarial opponent (in red) in the multi-agent environment YouShallNotPass. The adversary is trained via (\tb{top}) AP-MARL or (\tb{bottom}) IMAP. AP-MARL learns to statically collapse on the ground and fail to block the victim. In contrast, our IMAP learns a stronger adversarial skill to intercept the victim.}
	\label{fig: you-IMAP}
\end{figure}

\noindent\textbf{Summary of Contributions.} Our main contributions are summarized as follows:

\begin{itemize}[leftmargin=*]
	\item We propose IMAP—a general regularizer-based black-box AP learning method—and design four types of novel, well-motivated, and principled adversarial intrinsic regularizers, i.e., SC-, PC-, R-, and D-driven, in both single- and multi-agent environments.
	\item In single-agent environments, our IMAP outperforms the baseline SA-RL in four dense-reward locomotion tasks and nine sparse-reward tasks, including six locomotion, two navigation, and one manipulation tasks.
	\item In single-agent environments, our IMAP successfully evades two types of state-of-the-art defense methods, including adversarial training (e.g., ATLA and ATLA-SA~\cite{zhang2021robust}) and robust regularizer (e.g., SA~\cite{zhang2020robust}, RADIAL~\cite{oikarinen2021robust}, and WocaR~\cite{liang2022efficient}, shown in \Cref{fig: walker2d-IMAP}). We empirically show that a defense method that successfully defends one type of IMAP attack can fail against another type of IMAP, raising a new challenge for developing robust RL algorithms.
	\item In multi-agent environments, our IMAP achieves a state-of-the-art attacking success rate of 83.91\% in the competitive game YouShallNotPass, outperforming the baseline AP-MARL~\cite{gleave2019adversarial}. The adversary learns a natural blocking skill with the policy-coverage-driven adversarial intrinsic regularizer, shown in \Cref{fig: you-IMAP}. IMAP also outperforms AP-MARL in another competitive game, KickAndDefend.
     \item We develop a novel bias-reduction method for IMAP based on the adversarial optimality constraint and empirically demonstrate that it can effectively boost performance in both single- and multi-agent environments.
\end{itemize}

\section{Preliminaries}
\label{sec: preliminaries}

We introduce the formulations of single- and multi-agent RL tasks and the basic policy optimization method in this section. In all tasks, the goal of the victim is to maximize its expected episode rewards, while the adversary aims to minimize the expected episode rewards of the victim.

\paragraph{Single-Agent RL Tasks} In single-agent tasks, the agent interacts with the environment by taking sequential actions according to the observed state at each step. This process is usually modeled as a Markov Decision Process (MDP) $M=(\mathcal{S}, \mathcal{A}, P, R_E, \gamma, \mu)$. $\mathcal{S}$ and $\mathcal{A}$ are the state space and action space. $P: \mathcal{S} \times \mathcal{A} \to \Delta(\mathcal{S})$ is a transition function mapping state $s$ and action $a$ to the next state distribution $P(s'|s, a)$. $R_E: \mathcal{S} \times \mathcal{A} \times \mathcal{S} \to \mathbb{R}$ is the bounded extrinsic reward function. $\gamma \in [0,1)$ is the discount factor. And $\mu \in \Delta(\mathcal{S})$ is the initial state distribution.

\paragraph{Multi-Agent RL Tasks} For multi-agent tasks, we focus on two-player zero-sum competition games. A two-player zero-sum competition game can be formulated as a Markov Game $M=((\mathcal{S}^\bn,\mathcal{S}^\ra), (\mathcal{A}^\bn, \mathcal{A}^\ra), P, (R_E, -R_E), \gamma, \mu)$. $\mathcal{S}$ and $\mathcal{A}$ stand for the state and action space repsectively. Here, we use $\ra$ to represent the adversary and $\bn$ the victim. $P: \mathcal{S}^\bn \times \mathcal{S}^\ra \times \mathcal{A}^\bn \times \mathcal{A}^\ra \to \Delta(\mathcal{S}^\bn, \mathcal{S}^\ra)$ is the transition funtion where $\Delta(\mathcal{S}^\bn, \mathcal{S}^\ra)$ is the probability distribution space over $\mathcal{S}^\bn$ and $\mathcal{S}^\ra$. $R_E: \mathcal{S}^\bn \times \mathcal{S}^\ra \times \mathcal{A}^\bn \times \mathcal{A}^\ra \times \mathcal{S}^\bn \times \mathcal{S}^\ra \to \mathbb{R}$ is the bounded instant extrinsic reward function for the victim policy, and $-R_E$ is the corresponding extrinsic reward function for the adversarial agent according to the zero-sum assumption. $\gamma \in [0,1)$ is the common discount factor determining the horizon of the Markov Game, and $\mu \in \Delta(\mathcal{S}^\bn,\mathcal{S}^\ra)$ is the initial state distribution.

\paragraph{Policy Optimization} We use Proximal Policy Optimization (PPO)~\cite{schulman2017proximal} for AP learning. The objective function of PPO is defined as:
\begin{equation}
\label{eqn: PPO}
\begin{aligned}
	J^\text{PPO}(\pi) = \mathbb{E}_{s,a} \min  \left\{ \frac{\pi(a|s)}{\pi_k(a|s)} \hat{A}, \right.& \\
	\left. \operatorname{clip}\left( \frac{\pi(a|s)}{\pi_k(a|s)}; 1 - \epsilon, 1 + \epsilon \right)  \hat{A} \right\}&,
\end{aligned}
\end{equation}
where 
1) the density ratio $\frac{\pi(a|s)}{\pi_k(a|s)}$ is the importance weighting;
2) the clipping function $\operatorname{clip}(x;1-\epsilon, 1+\epsilon) = \begin{cases}
		1-\epsilon, &x\le 1 - \epsilon \\
		1 + \epsilon, &x\ge 1 - \epsilon \\
		x, & \text{otherwise}
		\end{cases}$
is to make sure that the policy gradient is zero when $| 1-\frac{\pi(a|s)}{\pi_k(a|s)} | \ge\epsilon$;
3) the advantage function $\hat{A}$ is estimated by Generalized Advantage Estimation (GAE)~\cite{schulman2015high} to reduce the variance of policy gradient estimation,
that is, $\hat{A}(s_t) = \sum_{l=0}^{\infty}(\gamma \lambda)^l (R_E(s_t,a_t,s_{t+1}) + \gamma V^{\pi_k}(s_{t+l+1}) - V^{\pi_k}(s_{t+l}))$; and
4) the outer minimization operator ensures the objective function $J^\text{PPO}$ is a lower bound of the objective $\mathbb{E}_{s,a} A$. Intuitively, this objective function makes sure the new and old policies are not so different.
PPO then utilizes multiple steps of mini-batch Stochastic Gradient Descent (SGD) on $J^\text{PPO}$ with a dataset $\mathcal{D}=\{(s, a,r_E,s')\}$ collected by the old policy $\pi_k$ and use regression to update the value function $V^{\pi}$.

\section{Threat Model}
\label{sec: threat model}

We adopt a black-box threat model for AP learning in both single- and multi-agent RL tasks. We describe the threat model from three aspects: objective, knowledge, and capabilities of the adversary.

\subsection{Objective of the Adversary}
\label{subsec: AP objective}

In both single- and multi-agent RL tasks, the goal of the attacker is to learn an optimal AP $\pi^\ra$ that can \emph{minimize} the test-time expected episode rewards of the deployed black-box victim policy $\pi^\bn$. We denote the adversarial state distribution induced by both $\pi^\ra$ and $\pi^\bn$ as $d^{\pi^\ra}=d^{\pi^\ra;\pi^\bn}$ to make the math notations concise since $\pi^\bn$ is held fixed. We define the test-time expected episode rewards of the victim policy as
\begin{equation}
	J^{\bn}_E(d^{\pi^\ra})= \sum_s d^{\pi^\ra} \hat{r}^{\bn}_E,
\end{equation}
where $\hat{r}^{\bn}_E$ is the surrogate reward of the victim policy since we assume the adversary cannot access the training-time reward $r^{\bn}_E$ of the victim policy. $r^{\bn}_E$ may contain complex reward shaping terms, while $\hat{r}^{\bn}_E$ is a simple indicator that the victim completes the task (e.g., runs far enough in locomotion or reaches the target position in navigation and manipulation) in the single-agent environment or win the competitive game in the multi-agent environment, that is, $\hat{r}^{\bn}_E = \mathbb{1}(\text{the victim succeeds})$.
The objective of the AP is then
\begin{equation}
\label{eqn: AP}
	J^\text{AP}(\pi^\ra) = -J^{\bn}_E(d^{\pi^\ra}).
\end{equation}

\subsection{Knowledge of the Adversary} In both single- and multi-agent RL tasks, the knowledge of the attacker is black-box, and the deployed victim policy network is assumed to be held fixed. Specifically, we assume that the adversary does not know the following information of the victim policy $\pi^\bn$: 1) training-time hyperparameters; 2) \emph{training-time rewards} $r^{\bn}_E$ and the \emph{value function} $V^{\pi^\bn}$; 3) test-time model architecture, parameters and activations.

\noindent\textbf{Clarification.} Here, we clarify the assumptions made above. The first and third assumptions are typical black-box assumptions adopted by all existing back-box AP learning methods, including SA-RL, AP-MARL, and Wu et al.'s method. For the second assumption, it is worth noting that only the victim policy network is utilized during the deployment phase in RL tasks. Therefore, for evasion attacks against RL, it is reasonable that both the training-time rewards $r^{\bn}_E$ and the value function $V^{\pi^\bn}$—which are only used in the training phase—are unknown to the adversary. SA-RL relaxed the second assumption, resulting in a weaker threat model.

\subsection{Capabilities of the Adversary}

Here, we introduce the adversary's capabilities separately in single- and multi-agent tasks since their transition functions are different, as stated in \Cref{sec: preliminaries}.

\paragraph{Single-Agent RL Tasks} The attacker can add small perturbations to the victim policy's inputs. We model the attacker as a state adversary $ \pi^\ra(\cdot|s)$, which can generate an adversarial perturbation $a^\ra \sim \pi^\ra$ based on the victim's current state $s^\bn$. The perturbation $a^\ra$ is bounded in an $\ell_p$ norm ball with a constant small radius $\epsilon$, that is, $\|a^\ra\|_p\le\epsilon$. The transition function under this threat model becomes $P^\ra(s^{\bn}_{t+1}|s^{\bn}_t,a^{\ra}_t) = P(s^{\bn}_{t+1}|s^{\bn}_t, \pi^\bn(s^{\bn}_t + a^{\ra}_t))$.

\paragraph{Multi-Agent RL Tasks} We focus on two-player zero-sum competitive games. The attacker can control an opponent agent $\ra$ to battle with the victim agent $\bn$, as visualized in \Cref{fig: you-IMAP}. Since the victim policy is held fixed, the two-player Markov game $M$ reduces to a single-player MDP $M^\ra = ((\mathcal{S}^\bn,\mathcal{S}^\ra), \mathcal{A}^\ra, P^\ra, (R_E,-R_E), \gamma, \mu)$. The transition function under this treat model becomes $P^\ra(s^{\ra}_{t+1}|s^{\ra}_t,a^{\ra}_t) = P(s^{\bn}_{t+1}, s^{\ra}_{t+1}|s^{\bn}_t, s^{\ra}_t, \pi^\bn(s^{\bn}_t, s^{\ra}_t), a^{\ra}_t)$. In each interaction step, the victim agent takes its action $\pi^\bn(s^{\bn}_t, s^{\ra}_t)$ based on the current environment state $(s^{\bn}_t, s^{\ra}_t)$, and the adversarial agent samples its action $a^{\ra}_t\sim\pi^\ra(\cdot|s^{\bn}_t, s^{\ra}_t)$ simultaneously.

\noindent\textbf{On the Adversary's Capabilities.} In sum, the attacker can obtain the environment's current state—$s^\bn$ in single-agent tasks and $(s^{\bn}_t, s^{\ra}_t)$ in multi-agent tasks—and maliciously influence the victim policy. In single-agent tasks, the adversary can directly inject perturbations $a^{\ra}_t$ to the inputs of the victim policy, i.e., $\pi^\bn(s^{\bn}_t + a^{\ra}_t)$; in multi-agent tasks, the adversary can indirectly influence the victim policy by generating adversarial observations $s^{\ra}_t$ with an opponent agent, i.e., $\pi^\bn(s^{\bn}_t, s^{\ra}_t)$.

\section{Proposed Attack}

In this section, we introduce the detailed techniques of IMAP. We start with the design of its regularizer-based optimization objective and the resulting RL problem for regularizer-based black-box AP learning. We then introduce four types of principled and well-motivated adversarial intrinsic regularizers as particular design cases. Following this, we derive the details of how to solve the policy optimization problem of IMAP. Finally, we introduce a novel bias-reduction mechanism for IMAP to relieve the potential distraction caused by adversarial intrinsic regularizers.

\subsection{Optimization Objective of IMAP}
\label{sub: the objective of IMAP}

Under the black-box threat model, as discussed in \Cref{sec: threat model}, maximizing the objective of the AP $J^\text{AP}(\pi^\ra)$ with trivial exploration methods like SA-RL and AP-MARL suffers from sample inefficiency and suboptimal solutions. To address these issues, we propose adversarial intrinsic regularizers, which intrinsically motivate the AP to explore novel states so as to uncover the potential vulnerabilities of the victim policy and learn stronger attacking skills. To make a trade-off between exploration (i.e., maximizing the adversarial intrinsic regularizer) and exploitation (i.e., maximizing the objective of the AP), we introduce a \emph{regularization} approach by incorporating the adversarial intrinsic regularizer $J_I(d^{\pi^\ra})$ into the objective of the AP $J^\text{AP}(\pi^\ra)$. The resulting optimization objective of IMAP is formulated as follows:
\begin{equation}
\label{eqn: regularized objective}
	J^\text{IMAP}(\pi^\ra) = J^\text{AP}(\pi^\ra) + \tau_k J_I(d^{\pi^\ra}),
\end{equation}
where $\tau_k$ represents the temperature parameter that determines the strength of the regularization.

It is worth noting that our formulation for the optimization objective of IMAP $J^\text{IMAP}(\pi^\ra)$ is general. The adversarial intrinsic regularizer $J_I(d^{\pi^\ra})$ can be a general function depending on the current adversarial state distribution $d^{\pi^\ra}$ and all past adversarial state distribution $\{d^{\pi_i^{\ra}}\}_{i=0}^{k}$. The adversarial intrinsic regularizer is designed to encourage the exploration of the AP in a principled manner. We present four types of adversarial intrinsic regularizers for IMAP as specific design cases in the following section.

Based on \Cref{eqn: regularized objective}, the resulting policy optimization problem of IMAP becomes
\begin{equation}
\label{eqn: optimization problem of IMAP}
	\max J^\text{IMAP}(\pi^\ra),\ \text{s.t.}\ \pi^\ra \in\arg\max J^\text{AP}(\pi^\ra).
\end{equation}
The constraint is necessary so as to ensure that, at convergence, the optimal AP  for $J^\text{IMAP}(\pi^\ra)$ is optimal for the objective of the adversary $J^\text{AP}(\pi^\ra)$. We name this constraint the adversarial optimality constraint.

\noindent\textbf{Uncovering Potential Vulnerabilities of the Victim Policy.} Before delving into the design of adversarial intrinsic regularizers, it is crucial to define the potential vulnerabilities of a victim policy. Formally, what we are looking for is a state region in the victim policy's state space, that is, $\mathcal{W}^\bn\in\mathcal{S}^\bn$, where $\hat{r}^{\bn}_E$ is small or zero. In other words, $\mathcal{W}^\bn$ is the state region that all sub-optimal trajectories of the victim policy pass through. Thus, uncovering the potential vulnerabilities of the victim policy entails diverting the victim policy from its optimal trajectories. This definition is consistent with the objective of the AP. In single-agent tasks, since $\mathcal{S}^\ra=\mathcal{S}^\bn\ni\mathcal{W}^\bn$ and $\|a^\ra\|_p\le\epsilon$, we can encourage the adversary to directly explore $\mathcal{S}^\ra$ to find $\mathcal{W}^\bn$. On the contrary, in multi-agent tasks, $\mathcal{S}^\bn\neq\mathcal{S}^\ra$, and the victim's and the adversary's states $s^\bn$ and $s^\ra$ are coupled by the transition function $P^\ra(s^{\ra}_{t+1}|s^{\ra}_t,a^{\ra}_t)$ derived in \Cref{sec: threat model}. Thus, we can design adversarial intrinsic regularizers in $\mathcal{S}^\ra$, $\mathcal{S}^\bn$, or $(\mathcal{S}^\ra,\mathcal{S}^\bn)$, to encourage the AP to uncover $\mathcal{W}^\bn$.
 
\subsection{Adversarial Intrinsic Regularizer Design}
\label{subsec: IO for IMAP}

We now introduce how to design appropriate adversarial intrinsic regularizers for black-box AP learning. Recall the objective of the adversary is to maximize the objective of the AP $J^\text{AP}(\pi^\ra)$. Existing black-box AP learning methods in both single-agent and multi-agent RL tasks typically rely on the heuristic exploration technique, which involves random perturbation on the outputs of the AP without considering the learning process of the AP. However, these methods have been shown to be sample-inefficient and are prone to converging towards suboptimal solutions due to premature exploitation, particularly in sparse-reward tasks. To overcome these limitations, we design four types of adversarial intrinsic regularizers to stimulate the exploration of the AP, including SC-driven, PC-driven, R-driven, and D-driven regularizers.

\subsubsection{State-Coverage-Driven Regularizer}

The first type of adversarial intrinsic regularizer we design is the State-Coverage-driven (SC) regularizer. The SC-driven regularizer aims to encourage the AP to maximize the adversarial SC by maximizing the entropy of the adversarial state distribution $d^{\pi^\ra}$. For instance, in a single-agent navigation RL task, the AP learned via IMAP-SC can disrupt the victim policy by enticing it to move randomly in the whole map. The SC-driven regularizer for single-agent tasks can be defined as follows:
\begin{equation}
	J_I^{\text{SC}}(d^{\pi^\ra}) = -{\textstyle\sum\nolimits_s} d^{\pi^\ra}\ln d^{\pi^\ra}.
\end{equation}

For multi-agent RL tasks, to uncover potential vulnerabilities $\mathcal{W}^\bn$ of the victim policy, we can 1) lure the victim policy into uniformly covering $\mathcal{S}^{\bn}$, and 2) encourage the adversary itself to uniformly cover $\mathcal{S}^{\ra}$. To accomplish this, we define the marginal state distribution $d^{\pi}_{\mathcal{Z}}(z)= (1-\gamma) \sum_{t=0}^{\infty} \gamma^t P(\Pi_\mathcal{Z} (s_t)=z|\mu, \pi)$, where $\Pi_\mathcal{Z}$ is an operator mapping the full state into a projection space $\mathcal{Z}$. The SC-driven regularizer for multi-agent tasks is then formulated as
\begin{equation}
\label{eqn: SC for multi-agent}
	J_I^{\text{SC-M}}(d^{\pi^\ra}) = (1-\xi)J_I^{\text{SC}}(d^{\pi^\ra}_{\mathcal{S}^\ra}) + \xi J_I^{\text{SC}}(d^{\pi^\ra}_{\mathcal{S}^\bn}),
\end{equation}
where $\xi$ is a constant for balancing the two sub-objectives.

\subsubsection{Policy-Coverage-Driven Regularizer}

Next, we introduce the Policy-Coverage-driven (PC) adversarial intrinsic regularizer, which aims to intrinsically motivate the adversary to divert the victim policy from its past (optimal) trajectories so as to uncover its potential vulnerabilities efficiently. We define the adversarial explored regions, or adversarial PC, as the sum of all historical adversarial state distributions $\rho^\ra= \sum_{i=1}^k d^{\pi^{\ra}_i}$. For single-agent tasks, we design the PC-driven adversarial intrinsic regularizer as follows:
\begin{equation}
	J_I^{\text{PC}}(d^{\pi^\ra}) = -{\textstyle\sum\nolimits_s} \rho^\ra\ln \rho^\ra.
\end{equation}
This regularizer can be regarded as the entropy of the adversarial PC. It encourages the adversary to visit novel regions where ${\rho^\ra}$ is small.

With the definition of the marginal state distribution in the previous section, we can design the novel PC-driven regularizer for multi-agent tasks
\begin{equation}
\label{eqn: PC for multi-agent}
	J_I^{\text{PC-M}}(d^{\pi^\ra}) = (1-\xi)J_I^{\text{PC}}(d^{\pi^\ra}_{\mathcal{S}^\ra}) + \xi J_I^{\text{PC}}(d^{\pi^\ra}_{\mathcal{S}^\bn}).
\end{equation}
Here, the first term encourages the adversary to visit novel states beyond the explored regions, while the second term aims to derail the victim from its optimal trajectories. The parameter $\xi$ is used to balance the two sub-objectives.

\subsubsection{Risk-Driven Regularizer}
\label{subsubsec: R}
Besides the SC- and PC-driven adversarial intrinsic regularizers, we propose a novel Risk-driven (R) adversarial intrinsic regularizer for black-box AP learning. The concept of the risk is inspired by safety RL~\cite{tessler2018reward}, where a cost function $c(s)$ is designed to constrain the behavior of the agent. For instance, when there exists a dangerous state $s^{\text{d}}$ in the state space, the cost function can be designed as $c(s) = -\|s-s^{\text{d}}\|$, penalizing the agent when it is close to $s^{\text{d}}$. By minimizing the expected cost function, the agent can be guided to stay away from $s^{\text{d}}$. In the context of evasion attacks, the attacker can maliciously select a potentially vulnerable state of the victim and lure the victim to approach this state. We refer to the state strategically selected by the adversary $\ra$ for the victim $\bn$ as the adversarial state $s^{\bn(\ra)}\in\mathcal{W}^\bn$. The corresponding cost function for the AP is then $c^\ra(s)=-\|\Pi_{\mathcal{S}^\bn} (s)-s^{\bn(\ra)}\|$. Here, we use the projector $\Pi_{\mathcal{S}^\bn} (s) \in \mathcal{S}^\bn$ to project the environment's full state $s$ into the victim policy's state space $\mathcal{S}^\bn$ since R only concerns the victim's states. The R-driven adversarial intrinsic regularizer for both single- and multi-agent tasks is then
\begin{equation}
	J_I^{\text{R}}(d^{\pi^\ra}) = -{\textstyle\sum\nolimits_s} d^{\pi^\ra}\|\Pi_{\mathcal{S}^\bn} (s)-s^{\bn;\ra}\|.
\end{equation}
Since all trajectories of the victim start from its initial state $s_0^{\bn}$, we have $s_0^{\bn}\in\mathcal{W}^{\bn}$. Thus, a natural choice of $s^{\bn(\ra)}$ is $s_0^{\bn}$.

\subsubsection{Divergence-Driven Regularizer}

We now introduce the fourth type of adversarial intrinsic regularizer, the D-driven adversarial intrinsic regularizer. The design of the D-driven regularizer is based on policy diversity~\cite{hong2018diversity} and~\cite{flet2021adversarially}. The objective of the D-driven regularizer is to intrinsically motivate the AP $\pi^\ra$ to continuously deviate from its past policies $\{\pi^{\ra}_i\}_{i=1}^{k}$, promoting diversity of the AP's behaviors and preventing the AP from being trapped in a local sub-optimal strategy. Note that we design the D-driven regularizer solely from the adversary's perspective, aiming to investigate whether this proprioceptive design can also help the AP discover potential vulnerabilities of the victim policy. Instead of randomly selecting an old policy from $\{\pi^{\ra}_i\}_{i=1}^{k}$, we introduce one adversarial mimic policy $\pi^{\ra;m}$ which has the same neural architecture as the AP $\pi^{\ra}$ and imitates the behaviors of these past policies $\{\pi^{\ra}_i\}_{i=1}^{k}$ by minimizing their average KL-divergence over all states, i.e., $\min\sum_s D_\text{KL} (\pi^{\ra;m}, \{\pi^{\ra}_i\}_{i=1}^{k})$. We then define the D-driven regularizer for both single- and multi-agent tasks as follows:
\begin{equation}
	J_I^{\text{D}}(d^{\pi^\ra}) = {\textstyle\sum\nolimits_s} d^{\pi^\ra} D_{\text{KL}}\left(\pi^{\ra}, \pi^{\ra;m} \right).
\end{equation}
By maximizing $J_I^{\text{D}}(d^{\pi^\ra})$, the AP  is encouraged to constantly deviate from its past policies to explore novel states in $\mathcal{S}^\bn$ to uncover $\mathcal{W}^\bn$ in a proprioceptive manner.

\noindent\textbf{Relationships Between the Four Types of Adversarial Intrinsic Regularizers.} Here, we clarify the relationships between the four types of adversarial intrinsic regularizers, i.e., SC-, PC-, R-, and D-driven adversarial intrinsic regularizers. They can be classified into two major categories, i.e., knowledge-based and data-based, depending on whether the regularizer involves only the agent's latest experiences (i.e., $d^{\pi^\ra}$) or the whole historical knowledge (i.e., $\{d^{\pi^{\ra}_i}\}_{i=1}^k$ or $\{\pi^{\ra}_i\}_{i=1}^k$). Thus, it is clear that SC- and R-driven regularizers belong to data-based since they only involve the adversary's latest state distribution $d^{\pi^\ra}$. In contrast, PC- and D-driven regularizers belong to knowledge-based because they both employ the adversary's all historical knowledge $\rho^\ra$ or $\{\pi^{\ra}_i\}_{i=1}^k$.

\noindent\textbf{State Density Approximation.}
To solve the optimization problem of IMAP, it is crucial to approximate the adversarial state density $d^{\pi^\ra}$ that all four regularizers we design involve. In the existing literature, there are two main types of methods for approximating state density, i.e., prediction-error-based and $K$-nearest-neighbour ($K$NN) estimation. Prediction-error-based methods, such as ICM~\cite{pathak2017curiosity} and RND~\cite{burda2018exploration}, directly estimate the inverse of state density using the prediction errors of a neural network. However, these methods suffer from forgetting problems~\cite {zhang2021noveld,zhang2021made}. We thus turn to the $K$NN method, a more efficient and stable nonparametric technique~\cite{liu2021behavior}. $K$NN estimates the state density via the inverse of the distance between a state and its $K$-nearest neighbor. Intuitively, the larger the distance, the smaller the state density (the sparser the samples). To estimate $d^{\pi^\ra}$, we cannot directly sample trajectories using the unsolved new policy $\pi^\ra$. In turn, we use the old policy $\pi^{\ra}_k$ to sample trajectories since PPO guarantees that $D_{\text{KL}}(P^{\pi^{\ra}_k} \| P^{\pi^\ra}) \leq \delta$. Thus, the estimated adversarial SC is given by $d^{\pi^\ra}(s) \approx 1/\|s-s^*_{\mathcal{D}_k}\|$. Here, $\mathcal{D}_k$ is a replay buffer containing trajectories sampled by only the latest old policy $\pi^{\ra}_k$, and $s^*_\mathcal{D}\in\mathcal{D}$ is the $K$-nearest state of $s$ in $\mathcal{D}$. Similarly, the adversarial PC can be estimated via $\rho^\ra(s) \approx 1/\|s-s^*_\mathcal{B}\|$, where $\mathcal{B}=\bigcup_{i=1}^k \mathcal{D}_i$ is the union replay buffer that contains all historical sampled trajectories. Note that one does not need to maintain the functional forms of all the old APs to estimate $\rho^\ra$. Instead, it is sufficient to sequentially store the trajectories sampled by the old policy $\pi^{\ra}_i$ at the $i$-th iteration of the policy optimization into $\mathcal{B}$ and use the replay buffer $\mathcal{B}$ to estimate the policy cover $\rho^\ra$ based on the $K$NN method.

\subsection{Solving the IMAP Optimization Problem}

We now present how to solve the policy optimization problem of IMAP defined in \Cref{eqn: optimization problem of IMAP}. It is easy to verify that $J^\text{IMAP}(\pi^\ra)$ is a concave function of $d^{\pi^\ra}$. Thus, we can leverage the Frank-Wolfe algorithm to solve this problem. Specifically, it iteratively solves the following problem
\begin{equation}
\label{eqn: frank-wolfe}
	\pi^{\ra}_{k+1} \in \arg\max \left\langle d^{\pi^\ra}, \nabla J^\text{IMAP}(\pi^{\ra}_k) \right\rangle
\end{equation}
to constructs a sequence of $\pi^{\ra}_0,\pi^{\ra}_1,...$ that converges to an optimal AP $\pi^{\ra *}$. The right-hand side of \Cref{eqn: frank-wolfe} is also known as the Frank-Wolfe gap~\cite{frank1956algorithm}. Maximizing the Frank-Wolfe gap is equivalent to finding a policy $\pi^{\ra}$ that maximizes the expected episode rewards, which is in proportion to $\nabla J^\text{IMAP}(\pi^{\ra}_k)$. Hence, we can obtain the adversarial intrinsic bonus as follows:
\begin{equation}
\label{eqn: intrinsic reward}
	r^{\ra}_I = \nabla J_I(d^{\pi^\ra}),
\end{equation}
and can derive the objective of IMAP based on \Cref{eqn: PPO}
\begin{equation}
\label{eqn: IMAP}
\begin{aligned}
	J^{\text{IMAP}}(\pi^\ra) = \mathbb{E}_{s,a} \min \left\{ \frac{\pi^\ra(a|s)}{\pi^{\ra}_k(a|s)} \left( \hat{A}_E+\tau_k \hat{A}_I \right), \right.& \\
	\quad\left. \operatorname{clip}\left( \frac{\pi^\ra(a|s)}{\pi^{\ra}_k(a|s)}; 1 - \epsilon, 1 + \epsilon \right) \left( \hat{A}_E+\tau_k \hat{A}_I \right) \right\},&
\end{aligned}
\end{equation}
where $\hat{A}_E$ and $\hat{A}_I$ are the estimated extrinsic and intrinsic advantage functions.

\begin{algorithm}[t]
\caption{IMAP}
\label{alg: IMAP}
\begin{algorithmic}
\STATE Initialize the AP $\pi^\ra$
\STATE Initialize replay buffers $\mathcal{B}$ and $\mathcal{D}$
\STATE Initialize counters $t=0$ and $k=0$
\STATE Initialize the temperature parameter $\tau_0=1$
\STATE Choose an adversarial intrinsic regularizer $J_I(d^{\pi^\ra})$
\WHILE {$t<T$}
\STATE {\color{lightgray_for_algorithm}\# \textit{Sampling Stage}}
\STATE Collect $\mathcal{D}=\{(s,a,-\hat{r}^{\bn}_E,s')\}$ using $\pi^{\ra}_k$ against $\pi^\bn$
\STATE Update the replay buffer $\mathcal{B} = \mathcal{B} \cup \mathcal{D}$
\STATE Update the sample counter $t = t + \operatorname{len}(\mathcal{D})$
\STATE {\color{lightgray_for_algorithm}\# \textit{Optimizing Stage}}
\STATE Compute the intrinsic bonus $r_I^{\ra}$ via \Cref{eqn: intrinsic reward}
\STATE Estimate advantages $\hat{A}_E$ and $\hat{A}_I$ via GAE
\STATE Update the AP $\pi^\ra$ via \Cref{eqn: IMAP}
\STATE Update value functions $V^{\ra}_E$ and $V^{\ra}_I$ via regression
\IF{use BR}
	\STATE Update $\tau_k$ via \Cref{eqn: tau} and \Cref{eqn: updating}
\ENDIF
\STATE Update the iteration counter $k=k+1$
\ENDWHILE
\end{algorithmic}
\end{algorithm}

\subsection{Reducing Bias in IMAP}

Though adversarial intrinsic regularizers can intrinsically motivate the AP to uncover the potential vulnerabilities of the victim policy, they may introduce bias to the optimal AP. In other words, the adversarial optimality constraint in \Cref{eqn: optimization problem of IMAP} may not hold, i.e., $\arg\max J^{\text{IMAP}}(\pi^\ra)\neq\arg\max J^{\text{AP}}(\pi^\ra)$. One common practice to reduce this bias is to perform a hyperparameter search to find the best sequences of the temperature parameter $\{\tau_i\}_{i=0}^T$ for different tasks. However, an exhaustive hyperparameter search is computationally expensive and sample-intensive.

\noindent\textbf{On Hyperparameter Search.}
Task-dependent temperature schedulers commonly utilize hyperparameter search to generate a sequence of the temperature parameter $\{\tau_i\}_{i=0}^T$ in advance, e.g., the exponentially decreasing scheduler $\tau_k=\beta(1-\rho)^k$ where both $\beta$ and $\rho$ are the hyperparameters to control the shape of the exponential function. Determining the optimal hyperparameters requires an expensive grid search. As the number of hyperparameters increases, the cost of the hyperparameter search grows exponentially. Conversely, our BR is a task-independent self-adaptive temperature scheduler that contains only one hyperparameter.

To address this challenge, we propose a novel adaptive Bias-Reduction (BR) method to ensure the adversarial optimality constraint. It is essential to balance the extrinsic objective and the intrinsic regularizer to ensure the maximization of the extrinsic objective (i.e., meeting the adversarial optimality constraint) rather than prioritizing the intrinsic regularizer at the end of the training process.
Specifically, we propose an approximate adversarial optimality constraint, that is,
\begin{equation}
\label{eqn: soft-constrained RL}
\begin{aligned}
	&\hspace{-1mm}\max J^\text{AP}(\pi^\ra) + J_I(d^{\pi^\ra}) \\
	&\text{s.t.}\ J^{\text{AP}}(\pi^\ra) >= J^{\text{AP}}(\pi^{\ra}_{k}).
\end{aligned}
\end{equation}
Once the approximate adversarial optimality constraint is satisfied, we have $J^{\text{AP}}(\pi^{\ra}_{k+1})\ge J^{\text{AP}}(\pi^{\ra}_{k})$, that is, the objective of the AP $J^{\text{AP}}$ monotonically increases.

To solve this soft-constrained optimization problem, we leverage the Lagrangian method to convert it into an unconstrained min-max optimization problem. The Lagrangian of \Cref{eqn: soft-constrained RL} is $\mathcal{L}(\pi^\ra, \lambda) = J^\text{AP}(\pi^\ra) + J_I(d^{\pi^\ra}) + \lambda( J^\text{AP}(\pi^\ra) - J^{\text{AP}}(\pi^{\ra}_{k})) \propto J^\text{AP}(\pi^\ra) + (1+\lambda)^{-1} J_I(d^{\pi^\ra})$ where $\lambda$ is the Lagrangian multiplier, and the corresponding dual problem is $\min_{\lambda\ge 0}\max_{\pi^\ra} \mathcal{L}(\pi^\ra, \lambda)$. By defining the temperature parameter $\tau_k$ as
\begin{equation}
\label{eqn: tau}
	\tau_k = (1 + \lambda_k)^{-1},
\end{equation}
we have $J^\text{IMAP}(\pi^\ra) = \mathcal{L}(\pi^\ra, \lambda_k)$. We alternatively update $\pi^\ra$ and $\lambda$, that is,
\begin{equation}
\label{eqn: updating}
\begin{aligned}
	\pi^{\ra}_{k+1} &\in\arg\max J^\text{IMAP}(\pi^\ra) \\
	\lambda_{k+1} &= \lambda_k - \eta (J^\text{AP}(\pi^{\ra}_{k+1}) - J^{\text{AP}}(\pi^{\ra}_{k})),
\end{aligned}
\end{equation}
to ensure that $J^\text{IMAP}$ and $J^\text{AP}$ are monotonically increased.

The form of the Lagrangian implies an interpretation for balancing the objective of the AP $J^\text{AP}$ and the adversarial intrinsic regularizer $J_I$. At the beginning of training, $\lambda_0=0$ and $\tau_0=1$, the AP focuses on exploring novel states to discover the potential vulnerabilities of the victim policy $\mathcal{W}^\bn$ via maximizing the sum of the objective of the AP $J^\text{AP}$ and the adversarial intrinsic regularizer $J_I$. When $\lambda$ grows as the training progresses, the AP pays more attention to exploiting the uncovered states in $\mathcal{W}^\bn$ via directly maximizing $J^\text{AP}$.

\section{Experiments}

We conduct comprehensive experiments in various types of RL tasks to evaluate our IMAP's attacking capacity and generalization with four types of adversarial intrinsic regularizers and verify the effectiveness of our bias-reduction method.

\subsection{Task Descriptions}
\label{subsec: task descriptions}

\begin{figure}[t]
	\centering
	\subcaptionbox{\label{fig: env-a}}{\includegraphics[width=0.24\columnwidth]{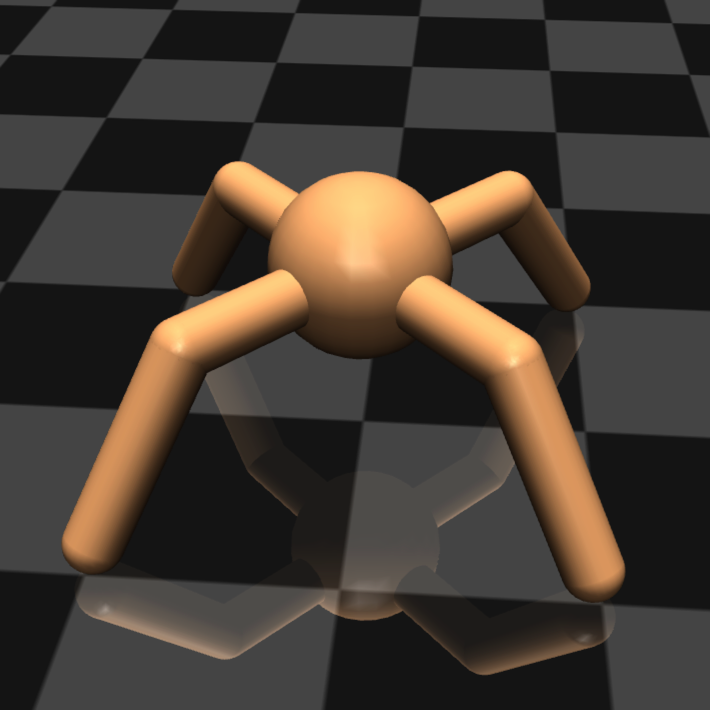}}
	\subcaptionbox{\label{fig: env-b}}{\includegraphics[width=0.24\columnwidth]{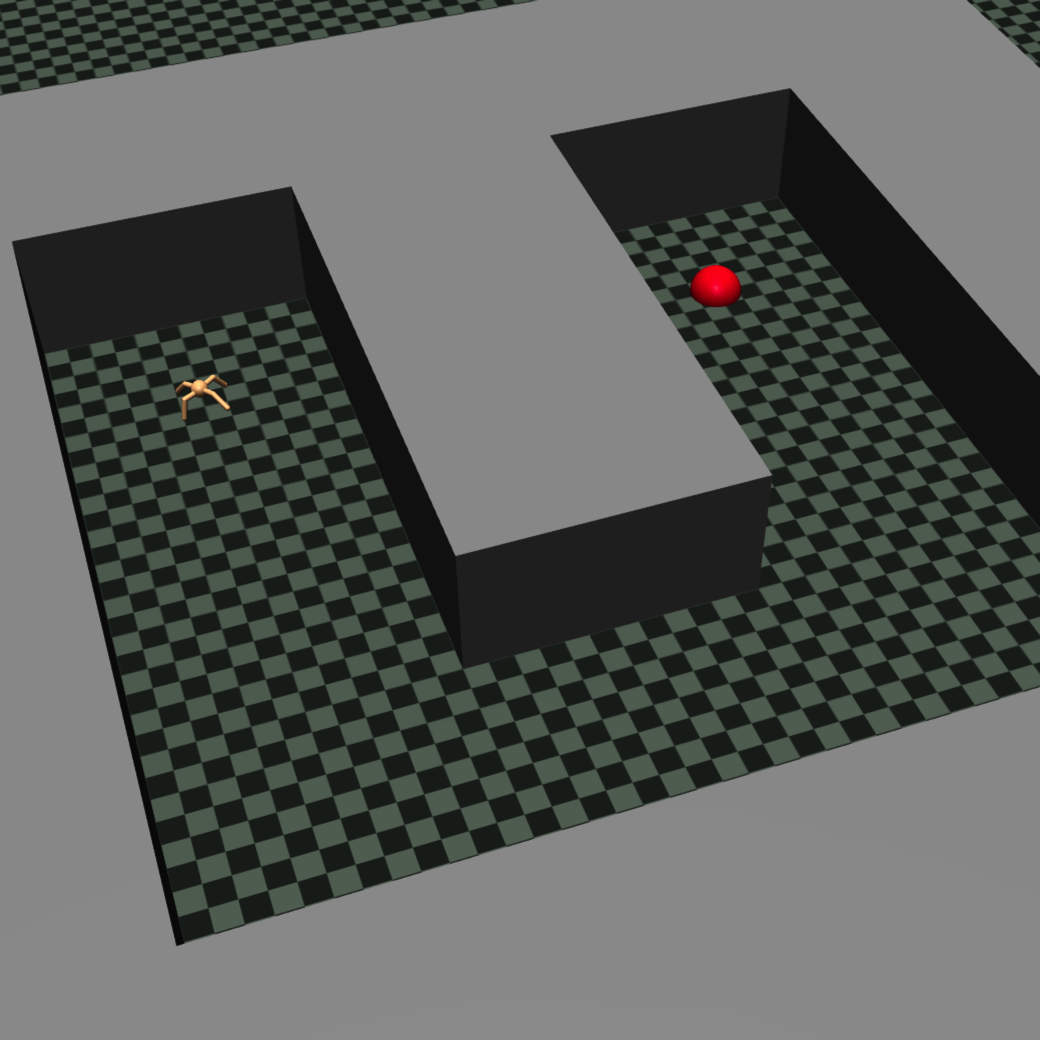}}
	\subcaptionbox{\label{fig: env-c}}{\includegraphics[width=0.24\columnwidth]{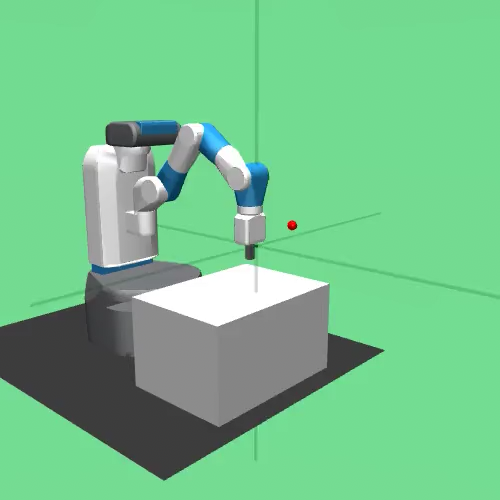}}
	\subcaptionbox{\label{fig: env-d}}{\includegraphics[width=0.24\columnwidth]{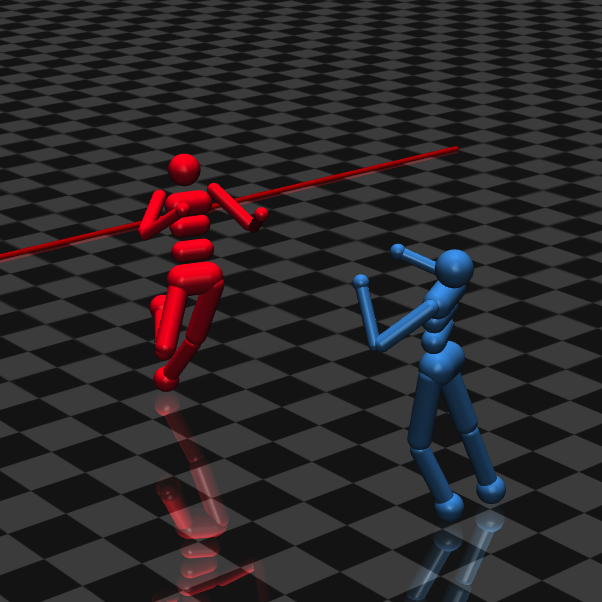}}
	\caption{Rendered pictures of typical MuJoCo environments. \subref{fig: env-a} the locomotion environment Ant; \subref{fig: env-b} the navigation environment AntUMaze where the red point is the goal position; \subref{fig: env-c} the manipulation environment FetchReach where the red point is the goal position; \subref{fig: env-d} the two-player zero-sum competitive game YouShallNotPass where the blue human is the victim and the red is the adversary.}
	\label{fig: env}
\end{figure}

In this section, we describe the details of the selected tasks. We evaluate our IMAP on both single- and multi-agent RL tasks. All environments are implemented based on the OpenAI Gym library and MuJoCo. For single-agent environments, we choose 1) four dense-reward locomotion tasks, including Hopper, Walker2d, HalfCheetah, and Ant~\cite{zhang2020robust,zhang2021robust,oikarinen2021robust,liang2022efficient}; 2) six sparse-reward locomotion tasks, including SparseHopper, SpasreWalker2d, SparseHalfCheetah, SparseAnt, SparseHumaonidStandup, and SparseHumanoid~\cite{mutti2021task,hazan2019provably}; 3) two sparse-reward navigation tasks, AntUMaze and Ant4Room~\cite{fu2020d4rl,eysenbach2022contrastive}; and 4) one sparse-reward manipulation task, FetchReach~\cite{plappert2018multi}. We choose two challenging two-player zero-sum competitive games, YouShallNotPass and KickAndDefend~\cite{bansal2017emergent,gleave2019adversarial,wu2021adversarial,guo2021adversarial,gong2022curiosity,li2023attacking}, as our multi-agent environments.

\paragraph{Criteria for Task Selection}
The selection of tasks in our experiments is based on two main criteria.
First, all tasks must be typical and have been adopted in former AP- and RL-related research works. This ensures that our evaluation is based on well-established benchmarks and allows for meaningful comparisons with existing methods.
Second, the types of tasks must be diverse to evaluate the attack capacity and generalization of our IMAP comprehensively.
In total, we selected 13 single-agent tasks and 2 multi-agent tasks that meet these criteria. Notably, tasks such as Ant, SpasreAnt, and YouShallNotPass have been used in multiple attacking and defense methods, making them suitable for comparative evaluations.
The selected single-agent tasks cover three types: locomotion, navigation, and manipulation. We specifically include a manipulation task to demonstrate that our IMAP can efficiently learn optimal black-box APs to attack agents in tasks other than locomotion tasks.
To further increase task diversity and evaluate IMAP's efficacy in multi-agent environments, we include two competitive games. These games involve victim agents with diverse skills, such as running and kicking.
Moreover, the dimension of the environment state varies across the single-agent tasks, ranging from 11 (Hopper) to 378 (Humanoid). In the multi-agent tasks, the dimension of the environment state grows to 378x2. This variation in the state space dimension allows us to assess the performance of IMAP across tasks with different levels of complexity.
Overall, our selected tasks ensure a comprehensive evaluation of IMAP.

\paragraph{Evaluation Metrics}
In our evaluation of single-agent tasks, we use the average episode rewards of the victim policy under attacks as the primary evaluation metric. This is a common metric used to assess the performance of the victim policy. A lower average episode reward indicates a more successful evasion attack, as the victim policy is less effective in achieving its intended goals.
For multi-agent tasks, we follow the previous works and report the attacking success rate of the AP. The attacking success rate is defined as $ASR = \frac{\text{\# of episodes where the adversary wins}}{\text{\# of total episodes}}$. It is easy to observe that $ASR = J^{\text{AP}}+1$. A higher $ASR$ indicates a stronger AP.

\paragraph{Single-Agent Tasks}
In dense-reward single-agent tasks, the victim agent is expected to run as fast as possible and live as long as possible.
According to the threat model in \Cref{sec: threat model}, the adversary cannot access the victim's training-time reward $r^{\bn}_E$ which contains complex reward shaping terms like $-\omega^{\bn}_a\|a^{\bn}\|^2$ and $- \omega^{\bn}_f\|f^{\bn}\|^2$. Instead, the adversary uses the surrogate reward $\hat{r}^{\bn}_E$.
In sparse-reward single-agent tasks, the victim agent is required to reach a certain goal at the end of the episode.
In four locomotion tasks, the victim agent starts from the initial position and must move forward across a distant line to complete the task.  The Ant environment is rendered in \Cref{fig: env-a}. The episode is terminated once the victim agent gets the extrinsic reward or enters an unhealthy state.
In two navigation tasks, the victim agent must navigate an Ant on different maps to reach a target region instead of always moving forward. This kind of task is thus known as more challenging than locomotion tasks like Ant and SparseAnt \cite{fu2020d4rl}. The environment AntUMaze is shown in \Cref{fig: env-b}.
In the manipulation task FetchReach, the robot arm is reset to an initial posture in each episode, and the victim agent is demanded to control the arm to move the end effector to a target position. FetchReach is visualized in \Cref{fig: env-c}.

\paragraph{Multi-Agent Tasks}
In YouShallNotPass, two humanoid robots are initialized facing each other. The victim policy controls the runner (in blue), while the AP controls the blocker (in red), as visualized in \Cref{fig: env-d}. The victim wins if it reaches the finish line within 500 timesteps, whereas the adversary wins if the victim does not. KickAndDefend is a soccer penalty shootout between two humanoid robots. The victim policy controls the kicker (in blue), and the AP controls the goalie (in red). The victim wins if it shoots the ball into the red gate; otherwise, the adversary wins. The victim policies were trained via self-playing against random old versions of their opponents.

\subsection{Baselines and Implementation}

We now introduce the baselines used in our experiments.

\paragraph{Single-Agent Tasks} We select SA-RL~\cite{zhang2021robust}, the state-of-the-art black-box AP learning method for single-agent tasks, as the baseline.
The original SA-RL relaxes the black-box assumption and requires the training-time reward $r^{\bn}_E$ to learn the optimal AP. To ensure a fair comparison, we implement both SA-RL and IMAP with the same simple surrogate reward $-\hat{r}^{\bn}_E$ defined in \Cref{subsec: AP objective} across all tasks. Moreover, all evasion attack methods for single-agent tasks in our experiments use the same attacking budget $\epsilon$ in each task.
To justify the choice of the baseline, here we discuss other related AP methods for single-agent tasks. Yu et al.'s method~\cite{yu2022natural} is tailored for video games. Sun et al.'s method~\cite{sun2021strongest} and Mo et al.'s method~\cite{mo2022attacking} fall under the category of white-box AP methods, as they necessitate access to the accurate model architecture and parameters of the victim policy. What is more, SA-RL outperforms MaxDiff and Robust Sarsa in their original paper~\cite{zhang2021robust}. Thus, SA-RL is the most suitable choice for our baseline in single-agent tasks.

\paragraph{Multi-Agent Tasks} We choose AP-MARL~\cite{gleave2019adversarial} as the baseline, which is recognized as the state-of-the-art black-box AP learning method for multi-agent tasks.
To justify this choice, we mention here other existing AP methods for multi-agent tasks. As highlighted in \Cref{sec: threat model}, our threat model is the same as that of AP-MARL.
Wu et al.'s method~\cite{wu2021adversarial}, while adopting the same threat model, introduces the requirement of training an extra surrogate victim model. This added complexity, however, results in only a marginal improvement when compared to AP-MARL. As reported in their original paper, Wu et al.'s method achieves an $ASR$ of only 60\% in YouShallNotPass, while AP-MARL achieves an $ASR$ of 59\% in our experiments.
Gong et al.'s method~\cite{gong2022curiosity} demands access to the training-time value function $V^{\pi^\bn}$ of the victim policy, thereby violating our threat model. In addition, Gong et al.'s method reports an $ASR$ of only 76\% in YouShallNotPass. In contrast, our method, IMAP-PC+BR, achieves a substantially higher $ASR$ of 83.91\% in the same environment without any relaxation of the black-box assumptions.
Guo et al.'s method~\cite{guo2021adversarial} extends AP-MARL to non-zero-sum competitive games and is the same as AP-MARL in zero-sum competitive games. Thus, AP-MARL is the ideal baseline for two-player zero-sum competitive games, which are our primary focus.

\subsection{Evaluation Results}

In this section, We report our main results. At a high level, our experiments reveal the following set of observations:
\begin{description}[leftmargin=*]
	\item[IMAP vs. SA-RL:] IMAP dominates SA-RL against most (15 out of 22) models and is comparable in the reset across all dense-reward single-agent tasks. Among all types of IMAP attacks, IMAP-PC achieves the best average performance.
	\item[Generalization:] IMAP excels in terms of generalization, surpassing SA-RL across diverse types of tasks, including locomotion, navigation, and manipulation tasks.
	\item[Choice of Adversarial Intrinsic Regularizers:] IMAP-PC is a suitable choice for a novel task since it exhibits superior generalization across our proposed four types of IMAP attack.
	\item[Effect of BR in IMAP:] The use of the balancing method BR in IMAP proves effective in enhancing the attacking performance, particularly when the adversarial intrinsic bonuses strongly distract the adversary.
	\item[IMAP vs. AP-MARL:] IMAP-PC+BR significantly outperforms AP-MARL in two zero-sum competitive games.
	\item[Hyperparameter Sensitivity:] IMAP displays resilience to variations in two newly introduced hyperparameters within reasonable bounds, i.e., $\xi$ in \Cref{eqn: PC for multi-agent} and $\eta$ in \Cref{eqn: updating}.
	\item[Evading Defense Methods:] IMAP successfully evades two different types of robust training defense methods, namely, adversarial training and robust regularizer.
\end{description}

\subsubsection{Performance in Dense-Reward Tasks} We first discuss the results of IMAP v.s. SA-RL in dense-reward tasks shown in \Cref{tab: results in dense-reward tasks}.

\begin{table*}[t]
	\centering
	\caption{Average episode rewards $J^{\bn}_E$ $\pm$ standard deviation of one vanilla model trained via PPO and five robust models trained using various defense methods over 300 episodes in four dense-reward locomotion tasks under no attack, random attack, SA-RL, and our four types of IMAP attacks. We \tb{bold} the best attack result (the lowest value) in each row and also report the average Attack Performance across all models. IMAP—the best of the four types of IMAP attacks—outperforms SA-RL against most (15 out of 22) models and exhibits similar performance in the reset across all dense-reward single-agent tasks. Among all attacks, IMAP-PC performs the best regarding the average performance.}
	\label{tab: results in dense-reward tasks}
	\begin{tabular}{wl{2cm}wc{\tabwidth}wc{\tabwidth}wc{\tabwidth}wc{\tabwidth}wc{\tabwidth}wc{\tabwidth}wc{\tabwidth}wc{\tabwidth}}
		\toprule
		\tb{Env.} & \tb{Victim} & \tb{No Attack} & \tb{Random} & \tb{SA-RL} & \tb{IMAP-SC} & \tb{IMAP-PC} & \tb{IMAP-R} & \tb{IMAP-D}\\
		\midrule
		\multirow{6}{0.8cm}{\tb{Hopper} 11D 0.075}
		& PPO (va.) & 3167\p542 & 2101\p793 & \TB{80}\p\TB{2} & \TB{80}\p\TB{2} & \TB{80}\p\TB{2} & \TB{80}\p\TB{2} & \TB{80}\p\TB{2}\\
		\cmidrule{2-9}
		& ATLA      & 2559\p958 & 2153\p882 & 875\p145 & 689\p132 & \TB{639}\p\TB{48} & 672\p120 & 808\p170\\
		& SA        & 3705\p2 & 2710\p801 & 1826\p897 & \TB{1282}\p\TB{68} & 1346\p85 & 1714\p1176 & 2278\p1144\\
		& ATLA-SA	& 3291\p600 & 3165\p576 & 1585\p469 & 1685\p512 & \TB{1536}\p\TB{392} & 1807\p642 & 1823\p527\\
		& RADIAL    & 3740\p44 & 3729\p100 & \TB{1622}\p\TB{408} & 2194\p672 & 1647\p398 & 1871\p498 & 1895\p551\\
		& WocaR		& 3616\p99 & 3633\p30 & 1850\p530 & 2140\p612 & \TB{1646}\p\TB{337} & 2917\p495 & 1832\p493\\
		\rowcolor{lightgray}\multicolumn{2}{c}{\tb{Average Across Victims}} & 3346 & 2915 & 1306 & 1345 & \TB{1149} & 1510 & 1452\\
		\midrule
		\multirow{6}{0.8cm}{\tb{Walker} 17D 0.05}
		& PPO (va.)	& 4472\p635 & 3007\p1200 & 1253\p468 & 1002\p391 & \TB{895}\p\TB{450} & 2966\p956 & 947\p160\\
		\cmidrule{2-9}
		& ATLA		& 3138\p1061 & 3384\p1056 & 1163\p464 & 1035\p614 & \TB{991}\p\TB{500} & 1599\p742 & 1385\p590\\
		& SA		& 4487\p61 & 4465\p39 & 3927\p162 & 4196\p231 & \TB{3072}\p\TB{1304} & 4083\p155 & 3820\p39\\
		& ATLA-SA	& 3842\p475 & 3927\p368 & 3508\p66 & 3144\p995 & \TB{2868}\p\TB{1145} & 3620\p143 & 3469+650\\
		& RADIAL	& 5251\p12 & 5184\p42 & \TB{4376}\p\TB{1229} & 4562\p941 & 4377\p1147 & 4584\p1021 & 4474\p1187\\
		& WocaR		& 4156\p495 & 4244\p157 & 2871\p1153 & 3178\p1168 & 2874\p1085 & \TB{2740}\p\TB{1162} & 2859\p1078\\
		\rowcolor{lightgray}\multicolumn{2}{c}{\tb{Average Across Victims}} & \cellcolor{lightgray}4224 & 4035 & 2850 & 2853 & \TB{2513} & 3265 & 2826\\
		\midrule
		\multirow{6}{0.8cm}{\tb{HalfCheetah} 17D 0.15}
		& PPO (va.) & 7117\p98 & 5486\p1378 & \TB{0}\p\TB{0} & \TB{0}\p\TB{0} & \TB{0}\p\TB{0} & 56\p147 & \TB{0}\p\TB{0}\\
		\cmidrule{2-9}
		& ATLA		& 5417\p49 & 5388\p34 & \TB{1696}\p\TB{1352} & 2451\p1352 & 1711\p1357 & 1996\p965 & 1765\p1357\\
		& SA        & 3632\p20 & 3619\p18 & 2997\p22 & 2996\p24 & \TB{2984}\p\TB{20} & 3390\p62 & 3000\p27\\
		& ATLA-SA	& 6157\p852 & 6164\p603 & \TB{4170}\p\TB{664} & 4311\p412 & 4202\p726 & 4395\p728 & 4231\p681\\
		& RADIAL    & 4724\p14 & 4731\p42 & 1654\p1312 & 1669\p1326 & \TB{1641}\p\TB{1298} & 1791\p1278 & 2563\p1496\\
		& WocaR		& 6032\p68 & 5969\p149 & 4257\p1254 & \TB{3734}\p\TB{1512} & 4026\p1374 & 4782\p105 & 4759\p487 \\
		\rowcolor{lightgray}\multicolumn{2}{c}{\tb{Average Across Victims}} & 5513 & 5226 & 2462 & 2433 & \TB{2427} & 2730 & 2720\\
		\midrule
		\multirow{4}{0.8cm}{\tb{Ant} 111D 0.15}
		& PPO (va.)	& 5687\p758 & 5261\p1005 & 351\p110 & 310\p184 & 212\p244 & \TB{188}\p\TB{135} & 284\p195\\
		\cmidrule{2-9}
		& ATLA		& 4894\p123 & 4541\p691 & \TB{0}\p\TB{0} & 428\p63 & 70\p128 & 696\p24 & \TB{0}\p\TB{0}\\
		& SA        & 4292\p384 & 4986\p452 & 2698\p822 & 2720\p879 & \TB{2643}\p\TB{851} & 2722\p994 & 2746\p831\\
		& ATLA-SA	& 5359\p153 & 5366\p104 & 3125\p207 & 3228\p190 & 3156\p302 & \TB{2611}\p\TB{213} & 3125\p182\\
		\rowcolor{lightgray}\multicolumn{2}{c}{\tb{Average Across Victims}}	& 5058 & 5039 & 1544 & 1672 & \TB{1520} & 1554 & 1539\\
		\bottomrule
	\end{tabular}
\end{table*}

\paragraph{IMAP Outperforms SA-RL} As shown in \Cref{tab: results in dense-reward tasks}, IMAP performs the best against most (15 out of 22) models (bolded results in each row) and is comparable to SA-RL in the rest. Here are some points that need to be explained. Firstly, when attacking the vanilla PPO models, IMAP significantly outperforms SA-RL in Walker (895 vs. 1253) and Ant (188 vs. 351) and performs equally in Hopper (both 80) and HalfCheetah (both 0). This underscores that when the victim policy has evident vulnerabilities, both SA-RL, utilizing the ad-hoc dithering exploration method, and IMAP, employing principled adversarial intrinsic regularizers, can readily identify and exploit these vulnerabilities to disrupt the victim. However, when there are more subtle vulnerabilities that elude trivial exploration methods, IMAP remains capable of efficiently discovering such vulnerabilities and further compromising the performance of the victim policy. Secondly, it is reasonable that there is no big difference between the performance of IMAP and SA-RL against certain models (e.g., 7 comparable cases beyond the 15 of 22 outperforming cases), such as 4377 vs. 4376 against Walker RADIAL and 4202 vs. 4170 against HalfCheetah ATLA-SA. This can be attributed to three factors: 1) the victim agent's strong robustness, making its vulnerabilities difficult to detect even with adversarial intrinsic regularizers; 2) the potential distraction introduced by the adversarial intrinsic regularizers, which may divert the adversarial policy from maximizing its core objective; 3) the worst cases in these 7 comparable tasks are easier to uncover compared to the other 15 outperforming tasks, causing that both IMAP and SA-RL can successfully discover the worst cases in these tasks and exhibit similar performances.
Note that the distraction phenomenon is more apparent in sparse-reward tasks. We delve deeper into it in the following section.

\paragraph{Choice of Adversarial Intrinsic Regularizers}
~Black-box robustness evaluation of the RL agent is a trial-and-error process where the agent knows nothing about the black-box victim model, including the training method, as stated in \Cref{sec: threat model}. We thus recommend starting the evaluation of a new black-box victim agent with IMAP-PC as the first trial since the experiments show it behaves well on average. As evident from \Cref{tab: results in dense-reward tasks}, IMAP-PC demonstrates the best average performance among all types of IMAP attacks. It notably reduces the average performance of all victim models by 65.66\%, 40.52\%, 55.97\%, and 69.94\% in Hopper, Walker, HalfCheetah, and Ant, respectively. For a comprehensive assessment of the robustness of a black-box victim policy, it is reasonable to explore multiple types of IMAP attacks. An essential insight from the results in \Cref{tab: results in dense-reward tasks} is that the type of potential vulnerabilities of the victim policy is not tied to the training method of that policy. For instance, the vulnerabilities of the ATLA-SA model in Ant can be identified via IMAP-R (reducing performance from 5359 to 2611), while the vulnerabilities of the ATLA-SA model in Walker can be exposed through IMAP-PC (reducing performance from 3842 to 2868). This pattern holds for other victim policy training methods as well. Therefore, it is advisable to try all adversarial intrinsic regularizers to discover potential vulnerabilities of the victim policy thoroughly. Additionally, we do not recommend combining multiple adversarial intrinsic regularizers since they may violate each other and make the adversary struggle.

\paragraph{On the Large Standard Deviation in the Performance of AP Attacks} The presence of substantial variance in the performance of reinforcement learning algorithms is a well-acknowledged phenomenon. This variability primarily stems from the inherent variance in policy gradient estimation~\cite{schulman2017proximal,haarnoja2018soft}. Given that the objective of the AP is maximized by PPO, it is unsurprising that the results exhibit large standard deviations. It is noteworthy that this phenomenon of significant standard deviation is not exclusive to IMAP but has also been reported in the original papers of SA-RL~\cite{zhang2021robust} and AP-MARL~\cite{gleave2019adversarial}. Importantly, variances do not significantly affect the application of AP methods. In practice, attackers have the flexibility to train multiple APs using various seeds and select the best one to attack the victim.

\subsubsection{Performance in Sparse-Reward Tasks.}
\label{sec: sparse-reward locomotion}

\begin{figure}[t]
	\centering
	\begin{subcaptionblock}{\columnwidth}
		\centering
		\includegraphics[width=0.99\columnwidth]{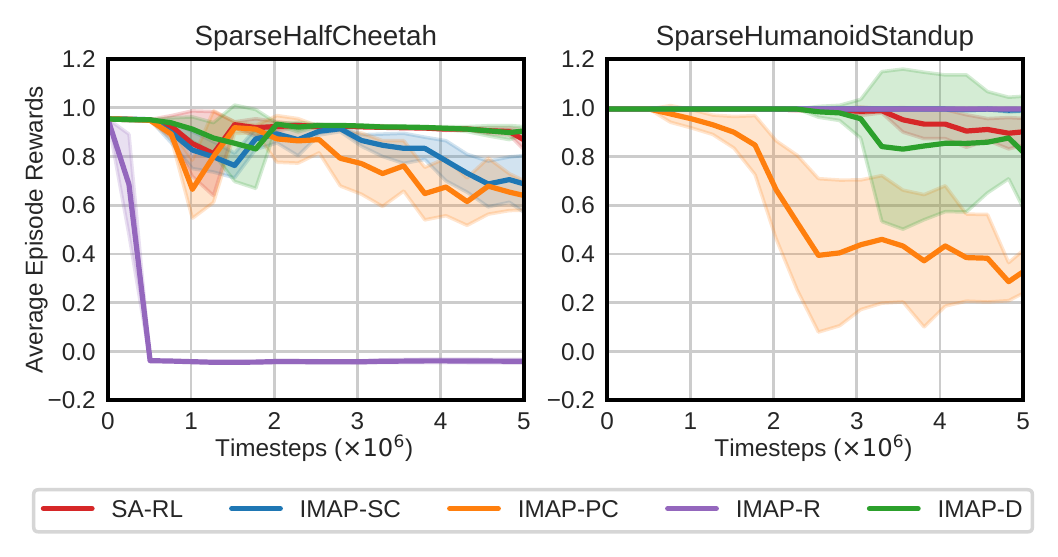}
	\end{subcaptionblock}
	\caption{Curves of test-time attacking results of SA-RL and four types of IMAP attacks on six sparse-reward locomotion tasks. IMAP-R significantly outperforms SA-RL in SparseHopper and SpareWalker2d; IMAP-PC significantly surpasses SA-RL in SparseHalfCheetah and SparseHumanoidStandup.}
	\label{fig: results in sparse-reward tasks}
\end{figure}

We now discuss the results in spares-reward single-agent tasks shown in \Cref{fig: results in sparse-reward tasks} and \Cref{tab: results in sparse-reward single-agent tasks}.

\begin{table*}[t]
	\centering
	\caption{Average episode rewards $J^{\bn}_E$ $\pm$ standard deviation of the victim policies over 1000 episodes across nine sparse-reward tasks, including six locomotion tasks (starting with 'S.'), two navigation tasks AntUMaze and Ant4Rooms, and one manipulation task, under nine attacks, including one baseline attack SA-RL, four types of IMAP attacks, and four types of IMAP+BR attacks. We \tb{bold} the best attack performance under each row. IMAP dominates SA-RL across all nine tasks (highlighted by \raisebox{0.18em}{\colorbox{lightgray}{\rule{0pt}{0.2em}\hspace{2em}}}). BR improves the attack performance of IMAP further in (4 out of 9) tasks.}
	\label{tab: results in sparse-reward single-agent tasks}
	\begin{tabular}{wl{2cm}wc{\secondtabwidth}wc{\secondtabwidth}wc{\secondtabwidth}wc{\secondtabwidth}wc{\secondtabwidth}wc{\secondtabwidth}wc{\secondtabwidth}wc{2.1cm}}
		\toprule
		\tb{Env.} & \tb{No Attack} & \tb{Random} & \tb{SA-RL} & \tb{IMAP-SC} & \tb{IMAP-PC} & \tb{IMAP-R} & \tb{IMAP-D} & \tb{IMAP+BR}\\
		\midrule
		S.Hopper & 0.95\p0.00 &0.95\p0.00 & 0.01\p0.32 & 0.00\p0.30 & 0.16\p0.45 & \cellcolor{lightgray}-0.03\p0.00 & -0.02\p0.28 & -\TB{0.05}\p\TB{0.22} (PC)\\
		S.Walker & 0.95\p0.00 & 0.94\p0.11 & 0.85\p0.23 & 0.66\p0.44 & 0.63\p0.45 & \cellcolor{lightgray}\TB{-0.04}\p\TB{0.01} & 0.91\p0.06 & 0.80\p0.32 (R)\\
		S.HalfCheetah & 0.98\p0.00 & 0.98\p0.00 & 0.30\p0.51 & 0.17\p0.45 & \cellcolor{lightgray}\TB{0.04}\p\TB{0.35} & 0.98\p0.00 & 0.33\p0.51 & 0.06\p0.37 (SC)\\
		S.Ant & 0.99\p0.00 & 0.98\p0.10 & 0.12\p0.42 & 0.23\p0.48 & 0.27\p0.49 & 0.43\p0.49 & \cellcolor{lightgray}0.12\p0.42 & \TB{0.10}\p\TB{0.40} (D)\\
		S.HumanStand & 0.99\p0.00 & 0.99\p0.00 & 0.88\p0.32 & 0.99\p0.05 & \cellcolor{lightgray}\TB{0.23}\p\TB{0.50} & 0.99\p0.00 & 0.80\p0.42 & 0.36\p0.54 (PC)\\
		S.Humanoid & 0.96\p0.00 & 0.93\p0.21 & 0.49\p0.50 & 0.46\p0.50 & 0.40\p0.49 & \cellcolor{lightgray}\TB{0.24}\p\TB{0.44} & 0.45\p0.5 & 0.35\p0.48 (PC)\\
		\midrule
		AntUMaze & 0.98\p0.00 & 0.98\p0.00 & 0.32\p0.52 & 0.30\p0.51 & 0.37\p0.52 & 0.97\p0.10 & \cellcolor{lightgray}0.28\p0.51 & \TB{0.19}\p\TB{0.47} (PC)\\
		Ant4Rooms & 0.91\p0.23 & 0.91\p0.00 & 0.34\p0.51 & 0.32\p0.51 & 0.40\p0.52 & 0.74\p0.43 & \cellcolor{lightgray}0.24\p0.48 & \TB{0.22}\p\TB{0.48} (R)\\
		\midrule
		FetchReach & 0.99\p0.00 & 0.99\p0.00 & 0.31\p0.50 & -0.10\p0.00 & \cellcolor{lightgray}\TB{-0.10}\p\TB{0.00} & 0.73\p0.42 & 0.51\p0.49 & \TB{-0.10}\p\TB{0.00} (PC)\\
		\midrule
		Average & 0.97 & 0.96 & 0.40 & 0.34 & \cellcolor{lightgray}0.28 & 0.56 & 0.40 & \TB{0.21}\\
		\bottomrule
	\end{tabular}
\end{table*}

\paragraph{Attacking Capacity and Generalization} The results presented in \Cref{tab: results in sparse-reward single-agent tasks} underscore IMAP's superior performance, outperforming SA-RL across all sparse-reward tasks. Additionally, as shown in \Cref{fig: results in sparse-reward tasks}, IMAP exhibits a significant advantage over SA-RL. In particular, SA-RL struggles to learn any effective attacking strategy with the trivial exploration method in SparseWalker2d. In contrast, IMAP-R efficiently discovers an optimal AP, leading to a remarkable reduction in the victim's average episode rewards, from 0.95 to -0.04, using only 0.5M samples (10$\times$ less than the 5M training sample budget). In SparseHumanoidStandup, SA-RL costs 5M samples to decrease the victim's performance from 0.99 to 0.88, while our IMAP-PC decreases the victim's performance to 0.4 within 2.5M samples (2$\times$ less).
In terms of generalization, IMAP consistently diminishes the performance of victim agents across all three types of tasks, including locomotion, navigation, and manipulation tasks. Moreover, IMAP surpasses SA-RL in terms of the average performance across tasks (in the last line of \Cref{tab: results in sparse-reward single-agent tasks}). These findings highlight the superior attacking capacity and generalization of IMAP compared to the baseline SA-RL.

\paragraph{Choice of Adversarial Intrinsic Regularizers} Again, we discuss the choice of the adversarial intrinsic regularizers in sparse-reward tasks. From \Cref{tab: results in sparse-reward single-agent tasks}, we observe that IMAP-PC mainly excels in locomotion and manipulation tasks; IMAP-D performs the best in navigation tasks; and IMAP-R stands out in partial locomotion tasks. These findings lead us to conclude that the suitability of an adversarial intrinsic regularizer is closely tied to the type of task. It is expected that different types of victim agents possess distinct potential vulnerabilities. For instance, in locomotion tasks like SparseHopper and SparseWalker, where the victim policy is dynamically unstable and prone to fall into unhealthy states under perturbations, the R-driven adversarial intrinsic regularizer is more likely to reveal these vulnerabilities. Considering that the average performance of IMAP-PC is the best, one may try IMAP-PC first and then other types of IMAP.

\paragraph{Effect of Bias-Reduction} The results presented in \Cref{tab: results in sparse-reward single-agent tasks} reveal that bias-reduction (BR) yields notable performance improvements for IMAP in several sparse-reward tasks.  Specifically, in Ant4Rooms, IMAP-R is heavily distracted by the R-driven regularizers. With the incorporation of BR, IMAP-R's performance is substantially enhanced, elevating it from 0.74 to 0.22. Note that 0.22$\pm$0.48 (R) indicates that this result is achieved by IMAP-R+BR. Similarly, IMAP-PC benefits from BR, improving its performance from 0.37 to 0.22 and emerging as the top-performing attack in AntUMaze. These outcomes underscore the efficacy of BR in augmenting the performance of IMAP in sparse-reward tasks.

\subsubsection{Performance in Competitive Games}
\label{sec: sparse-reward competition}

\begin{figure}[t]
	\centering
	\begin{subcaptionblock}{\columnwidth}
		\centering
		\includegraphics[width=0.99\columnwidth]{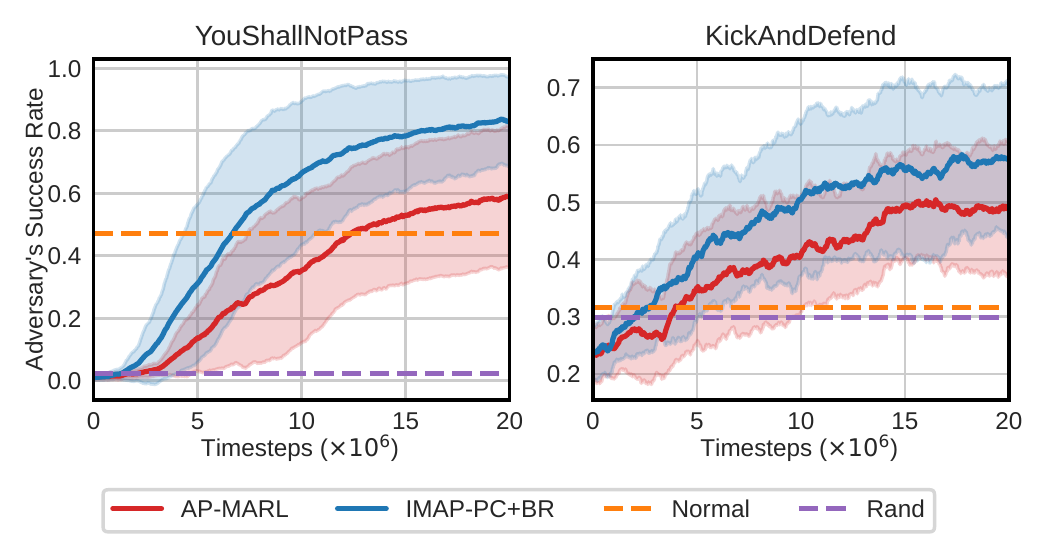}
	\end{subcaptionblock}
	\caption{Learning curves of AP-MARL and IMAP-PC+BR in two-player zero-sum competitive games. IMAP-PC+BR outperforms AP-MARL by a large margin.}
	\label{fig: results in multi-agent tasks}
\end{figure}

In this section, we discuss the results of IMAP v.s. AP-MARL in multi-agent tasks, as shown in \Cref{fig: results in multi-agent tasks}.

\paragraph{IMAP-PC+BR Outperforms AP-MARL} Building upon the insights gained from single-agent tasks, we delve into the performance of IMAP-PC+BR in two-player zero-sum competitive games in comparison to AP-MARL. Remarkably, IMAP-PC+BR consistently outperforms AP-MARL by a substantial margin.
As illustrated in \Cref{fig: results in multi-agent tasks}, IMAP-PC+BR consistently surpasses AP-MARL, substantially elevating the $ASR$ from 59.64\% to an impressive 83.91\%. This remarkable enhancement can be attributed to the acquisition of more natural attacking behavior in YouShallNotPass, as evidenced in \Cref{fig: you-IMAP}. In KickAndDefend, the game imposes constraints on the adversary (the goalie), confining it to a square region before the gate. Even within these constraints, IMAP manages to enhance the $ASR$ from 47.02\% to 56.96\%. These results reinforce the superior efficacy of IMAP-PC+BR in multi-agent tasks compared to AP-MARL, highlighting the effectiveness of the PC-driven regularizer in uncovering potential vulnerabilities in the victim policy.

\paragraph{Fundamental Reasons for Outperforming AP-MARL} The primary distinction lies in their exploration strategies employed during the training stage. AP-MARL utilizes a heuristic dithering exploration strategy, while IMAP+PC is intrinsically motivated by the PC-driven regularizer. The PC-driven regularizer allows IMAP to uncover the vulnerabilities of the victim $\mathcal{W}^\bn$ more efficiently through a larger coverage on the victim and the adversary's joint state space $(\mathcal{S}^\bn,\mathcal{S}^\ra)$.

\begin{figure}[t]
	\centering
	\begin{subcaptionblock}{\columnwidth}
		\centering
		\includegraphics[width=0.96\columnwidth]{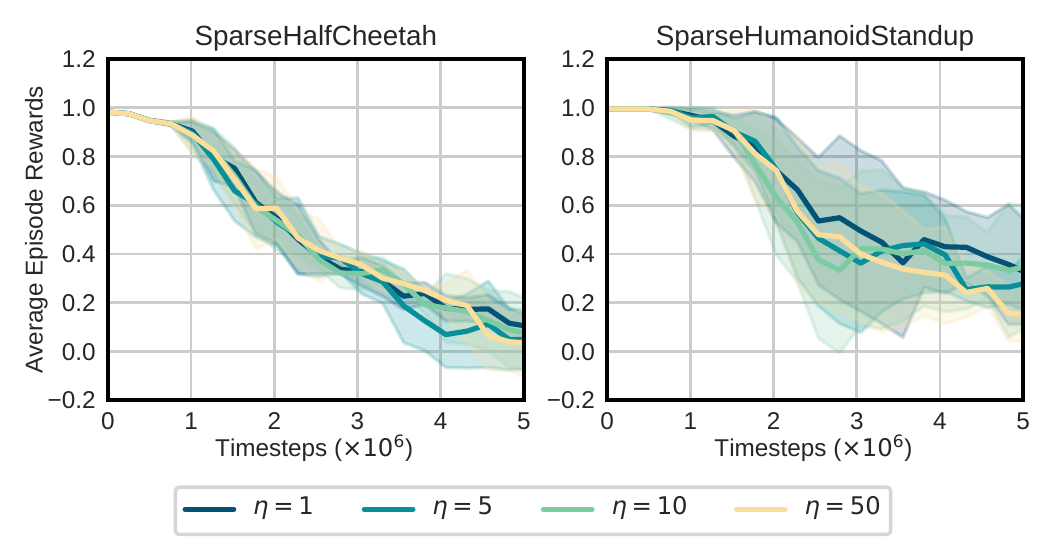}
	\end{subcaptionblock}
	\caption{Ablation study on the hyperparameter $\eta$ of IMAP.}
	\label{fig: ablation study on eta}
\end{figure}

\begin{figure}[t]
	\centering
	\begin{subcaptionblock}{\columnwidth}
		\centering
		\includegraphics[width=0.99\columnwidth]{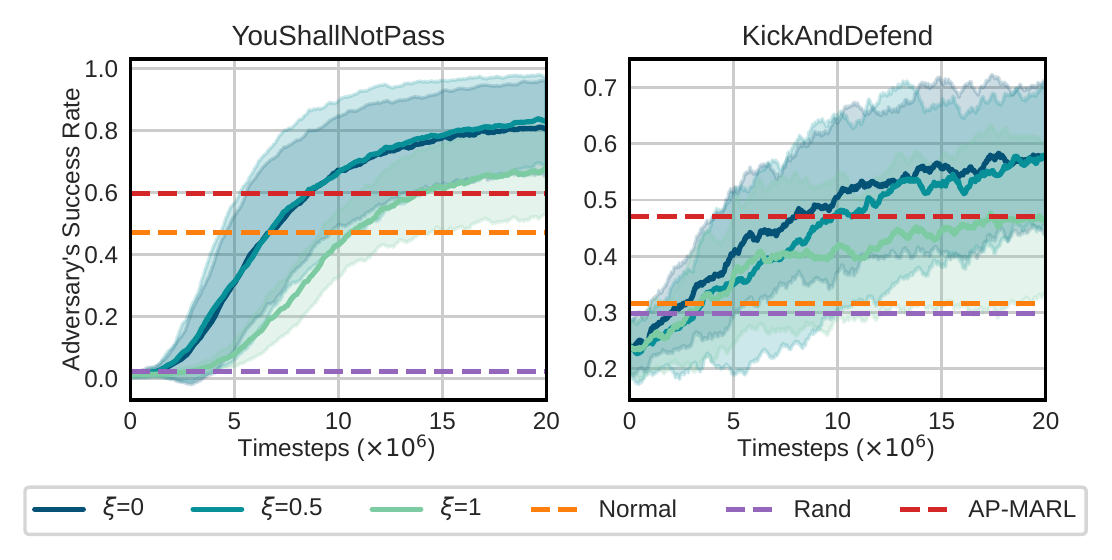}
	\end{subcaptionblock}
	\caption{Ablation study on the hyperparameter $\xi$ of IMAP.}
	\label{fig: ablation study on xi}
\end{figure}

\paragraph{Ablation Study on Hyperparameters} We conducted an in-depth investigation into the impact of IMAP's two newly introduced hyperparameters: the updating step size $\eta$ of the Lagrangian multiplier in \Cref{eqn: updating} and the constant $\xi$ for balancing the two sub-objectives in \Cref{eqn: PC for multi-agent}. \Cref{fig: ablation study on eta} and \Cref{fig: ablation study on xi} reveal the performance of IMAP-PC+BR under different hyperparameter settings in single- and multi-agent tasks separately. \Cref{fig: ablation study on eta} demonstrates that IMAP is insensitive to $\eta$ when $\eta\in\{1,5,10,50\}$. A larger updating step size leads to better performance within this range. \Cref{fig: ablation study on xi} shows that IMAP is also robust to changes of $\xi\in\{0.5,1\}$. Recall that $J_I^{\text{SC-M}}(d^{\pi^\ra}) = (1-\xi)J_I^{\text{SC}}(d^{\pi^\ra}_{\mathcal{S}^\ra}) + \xi J_I^{\text{SC}}(d^{\pi^\ra}_{\mathcal{S}^\bn})$. \Cref{fig: ablation study on xi} indicates that $J_I^{\text{SC}}(d^{\pi^\ra}_{\mathcal{S}^\ra})$ is critical for the performance of IMAP-PC. Note that when these hyperparameters go beyond rational ranges (i.e., [1,50] for the updating step size $\eta$ and (0,1] for the balancing constant $\xi$), the performance of IMAP may significantly deteriorate. For instance, when $\xi=0$, the ASR of IMAP drops from the optimum 83.91\% to the baseline ASR of 60\%.

\section{Defense Methods Against IMAP}

In this section, we explore potential defense methods against IMAP and evaluate IMAP's effectiveness against two main types of defense methods.

\noindent\textbf{Possible Defense Methods Against IMAP.} There are four categories of methods for RL agents to defend against evasion attacks: adversarial training, robust regularizer, randomized smoothing, and active detection. Adversarial training in the context of RL closely resembles its counterpart in DNN. It involves optimizing the policy under either gradient-based evasion attacks or the optimal AP. The adversary can have various access rights in the environment to robustify the victim agent against different types of uncertainties, e.g., directly injecting perturbations to the state or action or reward~\cite{behzadan2017whatever,vinitsky2020robust,tan2020robustifying,zhang2021robust,sun2021strongest,wu2022robust}, adding disturbance forces or torques~\cite{pinto2017robust}, or even changing the layout or dynamic property of the environment~\cite{chen2018gradient}. Robust regularizer aims to enhance the smoothness of the learned policy by upper-bounding the divergence of the action distributions under state perturbations~\cite{zhang2020robust,shen2020deep,oikarinen2021robust,liang2022efficient}. Randomized smoothing has been applied to analyze the robustness of reinforcement learning from a probabilistic perspective~\cite{anderson2022certified,kumar2021policy,wu2021crop,lutter2021robust}. Active detection strategies focus on identifying malicious samples by comparing the KL-divergence of the nominal action distribution and the predicted one~\cite{lin2017detecting} or using explainable AI techniques to identify critical time steps contributing to the victim agent’s performance~\cite{guo2021edge}.

\noindent\textbf{Evaluating IMAP Against Defense Methods.} There are two types of defense methods based on the above analysis, i.e., robust training (adversarial training and robust regularizer) and test-time defense mechanisms (randomized smoothing and active detection). We focus on the first type of defense method against IMAP and leave the second type of defense method for future work. What is more, randomized smoothing and active detection may sacrifice the victim's test-time performance since they operate on the original inputs of the deployed victim policy. Robust regularizer methods include 1) SA~\cite{zhang2020robust}, which improves the robustness of the victim agent via a smooth policy regularization (denoted as SA-regularizer for concision) on the victim policy solved by the convex relaxation technique; 2) RADIAL~\cite{oikarinen2021robust}, which leverages an adversarial loss function based on bounds of the victim policy under bounded $l_\infty$ attacks; and 3) WocaR~\cite{liang2022efficient}, which directly estimates and optimizes the worst-case episode rewards also based on bounds of the victim policy under bounded $l_\infty$ attacks. Two adversarial training methods include: 1) ATLA~\cite{zhang2021robust}, which alternately trains the victim agent and an RL attacker with independent value and policy networks; and 2) ATLA-SA~\cite{zhang2021robust}, which combines the training procedure of ATLA with the SA-regularizer and uses LSTM as the policy network. The results in \Cref{tab: results in dense-reward tasks} demonstrate our IMAP is effective in evading robust models trained by either adversarial training methods or robust regularizer methods. All victim models we adopt are publicly released. We report the average performance over 300 episodes to make the results statistically reliable. Notably, even against the state-of-the-art robust WocaR models, our IMAP can efficiently uncover their potential vulnerabilities via proper adversarial intrinsic regularizers under the black-box threat model, reducing their performance by 54.58\%, 34.07\%, and 38.10\% in Hopper, Walker, and HalfCheetah respectively.

\section{Discussion}

In this section, we provide an in-depth discussion of the sample efficiency of IMAP and identify the specific reinforcement learning engines or models that can benefit from the proposal of the IMAP.

\noindent\textbf{On the Sample Efficiency.}
There are three key insights on the sample efficiency of IMAP. Firstly, the adversarial intrinsic regularizers (i.e., SC, PC, R, D) contribute more to the sample efficiency of IMAP compared to BR. Intuitively, when the potential vulnerabilities of the victim policy are extremely difficult to discover, it becomes challenging to learn an optimal adversarial policy with an inappropriate or no intrinsic regulator. Secondly, there is a trade-off between sample efficiency and performance. As shown in \Cref{tab: results in sparse-reward single-agent tasks}, satisfactory results can be achieved by using the adversarial intrinsic regularizer PC alone, without the need for BR. Therefore, unless ultimate performance is sought, it is not necessary to increase the number of samples by 8x. Thirdly, IMAP-PC is based on policy cover theory that enjoys polynomial sample complexity~\cite{agarwal2020pc}. Intuitively, it is aware of the agent's entire historical knowledge and explicitly deviates the victim policy from its optimal trajectories. Hence, IMAP-PC is more likely to discover the victim's worst cases than SA-RL which explores randomly.

\noindent\textbf{RL Agents Benefiting From IMAP.}
There are various real-world scenarios for RL agents, e.g., Large Language Models (LLM)~\cite{brown2020language}, autonomous driving~\cite{kiran2021deep}, traffic control~\cite{wu2017flow}, industrial automation and manufacturing~\cite{oliff2020reinforcement}, dynamic treatment regimes~\cite{zhang2019near,zhang2020designing}, and recommendation systems~\cite{zou2019reinforcement,afsar2022reinforcement}. IMAP is promising to evaluate these deployed black-box real-world RL engines or models. Here, we provide two appropriate cases. Firstly, to evaluate the robustness of a real-world victim autonomous driving RL agent, we can use IMAP to either generate stealthy sensor noise to disrupt the victim car~\cite{zhang2020robust} or control another malicious car to intercept the victim car to make a traffic jam or even accident~\cite{sharif2022adversarial}. Secondly, we can formulate the red-teaming tasks for LLM as a two-player competitive game, regarding the target LLM as the victim agent and the red-teaming language model as the adversarial policy~\cite{hong2024curiositydriven}. In such a way, IMAP holds the potential for learning a strong, intrinsically motivated red-teaming adversarial policy to evaluate the robustness of the real-world commercial black-box LLM, e.g., GPT-4.

\section{Conclusion}

In this paper, we proposed a new regularizer-based AP learning method called IMAP to evaluate the robustness of test-time RL agents in single- and multi-agent environments under the black-box threat model. We presented four types of adversarial intrinsic regularizers that encourage the AP to explore novel states so as to uncover the potential vulnerabilities of the victim policy. We also introduced a novel balancing method, BR, to boost IMAP further. We conducted extensive evaluation experiments of IMAP across various types of tasks. The experimental results demonstrated that IMAP outperformed existing methods, including SA-RL and AP-MARL, in terms of attacking capability and generalization. We also empirically showed that BR effectively boosted IMAP in both single- and multi-agent environments.
Moreover, we demonstrated that IMAP successfully evaded state-of-the-art defense methods, including adversarial training and robust regularizer methods. Additionally, our ablation study showed IMAP was insensitive to its main hyperparameters. Note that though our proposed four adversarial intrinsic regularizers covered the main branches of intrinsic motivation, one can still design new adversarial intrinsic regularizers for IMAP as needed. We leave this as future work.

\section*{Acknowledgment}

We thank the anonymous reviewers and our shepherd, Dr. Stjepan Picek, for their helpful and valuable feedback, and Tencent Yunding Laboratory for providing computing resources and generous technical support.
This work was partially supported by HK RGC under Grants (CityU 11218322, R6021-20F, R1012-21, RFS2122-1S04, C2004-21G, C1029-22G, and N\_CityU139/21), the National Natural Science Foundation of China (U21B2018, 62161160337, 61822309, U20B2049, 61773310, U1736205, 61802166, 62276067), and Shaanxi Province Key Industry Innovation Program (2021ZDLGY01-02).

\bibliographystyle{myieeetr}
\bibliography{dsn2024ref}

\end{document}